\pdfoutput=1

\documentclass[11pt]{article}

\usepackage{EMNLP2022}
\usepackage{times}
\usepackage{latexsym}

\usepackage[T1]{fontenc}

\usepackage[utf8]{inputenc}

\usepackage{microtype}

\usepackage{inconsolata}

\usepackage{algorithm}
\usepackage{algorithmic}
\usepackage{amsmath}
\usepackage{graphicx}
\usepackage{booktabs}
\usepackage{multirow}
\usepackage{makecell}
\title{Stop Filtering: Multi-View Attribute-Enhanced Dialogue Learning}

\author{Yiwei Li\footnotemark[1], Bin Sun\footnotemark[1], Shaoxiong Feng\footnotemark[1], Kan Li\footnotemark[2]\\
  School of Computer Science, Beijing Institute of Technology \\
  \texttt{\{liyiwei,binsun,shaoxiongfeng,likan\}@bit.edu.cn}}
\begin{document}
\maketitle

\renewcommand{\thefootnote}{\fnsymbol{footnote}} 
\footnotetext[1]{Equal contribution.} 
\footnotetext[2]{Corresponding author.} 

\begin{abstract}
There is a growing interest in improving the conversational ability of models by filtering the raw dialogue corpora. Previous filtering strategies usually rely on a scoring method to assess and discard samples from one perspective, enabling the model to enhance the corresponding dialogue attributes (e.g., consistency) more easily. However, the discarded samples may obtain high scores in other perspectives and can provide regularization effects on the model learning, which causes the performance improvement to be sensitive to the filtering ratio. In this work, we propose a multi-view attribute-enhanced dialogue learning framework that strengthens the attribute-related features more robustly and comprehensively. Instead of filtering the raw dataset to train the model, our framework first pre-trains the model on the raw dataset and then fine-tunes it through adapters on the selected sub-sets, which also enhances certain attributes of responses but without suffering from the problems mentioned above. Considering the variety of the dialogue attribute, we further design a multi-view enhancement mechanism, including multi-view selection and inter-view fusion. It groups the high-quality samples from multiple perspectives, respectively, and enhances different attributes of responses with the corresponding sample sets and adapters, keeping knowledge independent and allowing flexible integration. Empirical results and analysis show that our framework can improve the performance significantly in terms of enhancing dialogue attributes and fusing view-specific knowledge.
\end{abstract}

\section{Introduction}

Neural dialogue generation \citep{Seq2Seq-Sordoni-2015,NoisyData-Vinyals-2015,Seq2Seq-ShangLifeng-2015} has gained increasing attention. Given the dialogue corpora, previous work focuses on how to improve the conversational ability of models by redesigning objectives \citep{MMI-LiJiwei-2016,VaeTextGeneration-Bowman-2016,SeqGan-YuLantao-2017} and network structures \citep{HRED-Dialogue-Serban-2016,HVaeMN-ChenHongshen-2018,DBLP:conf/acl/ZhangLPGC19} or introducing external knowledge \citep{FactKnowledge-Ghazvininejad-2018}. 
To facilitate the model learning, apart from that, it is also necessary to explore how to manipulate samples during training due to the \textit{noises} in the dialogue corpora. 
Recently, a line of work \citep{FilterCoherence-Xu-2018,FilterEntropy-Csaky-2019,FilterConsistency-Akama-2020} introduces a data manipulation strategy, called Data Filtering, to boost the model performance. 
Specifically, they first measure the quality of samples in terms of a certain dialogue attribute by a scoring method, and then discard the \textit{noisy} samples with low scores. The filtered data can induce the model to learn attribute-related features more effectively for the generation of high-quality responses. %

\begin{figure}[t]
\centering
\includegraphics[width=0.9\linewidth]{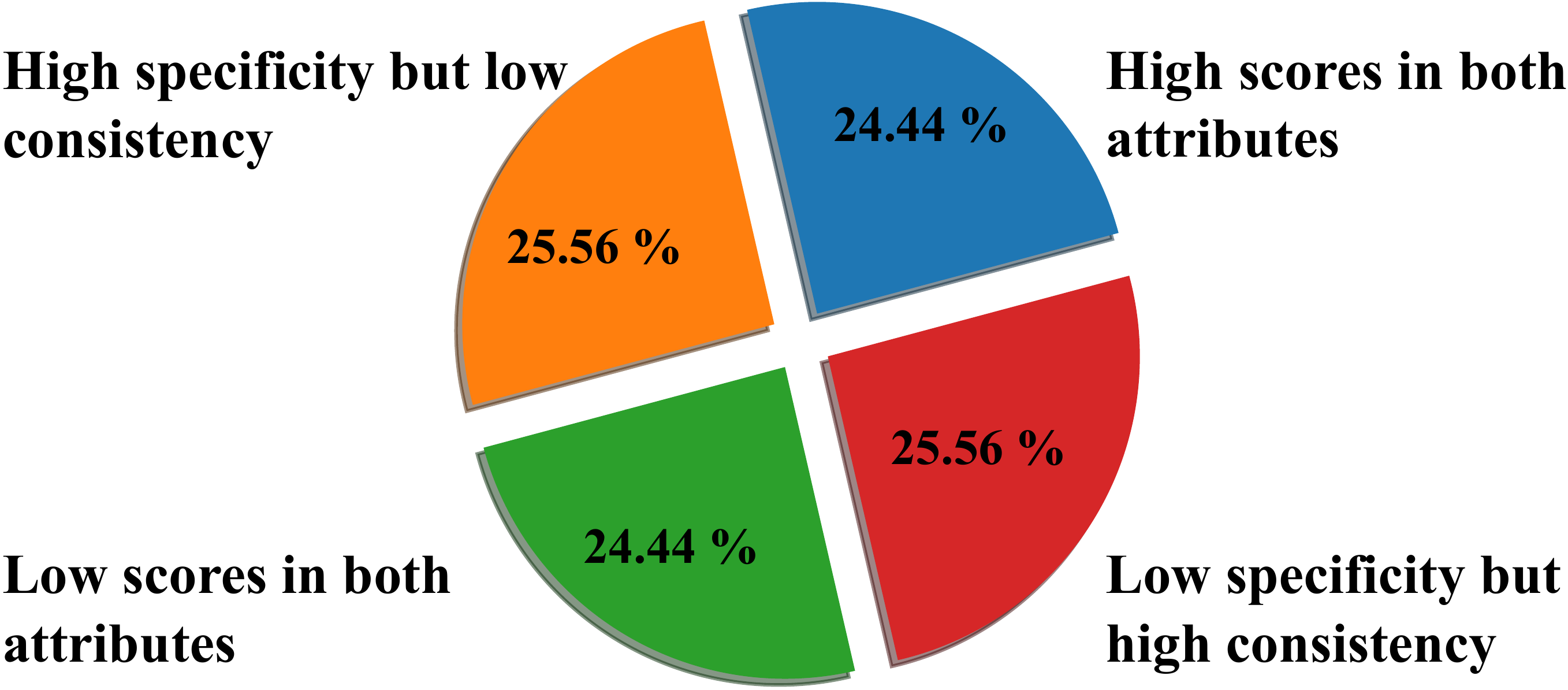}
\caption{The score distribution of training samples of DailyDialog in terms of \textit{Consistency} and \textit{Specificity}. Following \citet{FilterConsistency-Akama-2020}, we regard the top 50\% of the training set ranked by scores as high-quality for each attribute.}
\label{fig.multi_indicators}
\end{figure} 

However, the performance improvement from the data filtering strategy is sensitive to the filtering ratio, indicated by prior work \citep{FilterConsistency-Akama-2020} and our analysis in Section~\ref{data volume}.

It is because those discarded samples may obtain high scores in other perspectives and still benefit the feature learning of other attributes. 
We visualize the score distribution of training samples of DailyDialog \citep{dailydialog2017} in terms of two dialogue attributes, \textit{Consistency} \citep{FilterConsistency-Akama-2020} and \textit{Specificity} \citep{FilterSpecificity-See-2019}. As shown in Figure~\ref{fig.multi_indicators}, the samples in orange and red parts are regarded as high-quality in one perspective but low-quality in the others. 
Nevertheless, the experimental results (see Appendix~\ref{appendix:a}) show that after further considering either the orange or the red part, the model trained with the blue part before achieves better performance, demonstrating that samples with high scores in any meaningful perspective are high-quality. 
Therefore the filtering-based data manipulation strategy inevitably causes the model to neglect the learning of other dialogue attributes. 
Furthermore, the discarded samples, as the regularization factor, also prevent the model from overfitting to the filtered data consisting of fewer samples.

Another problem is that prior filtering-based work only improves the conversational ability from one perspective reflected by the proposed scoring method, which can not achieve the goal of the dialogue system, i.e., showing superiority in multiple perspectives simultaneously \citep{DBLP:journals/sigkdd/ChenLYT17}. 
A straightforward method is to use the union of different training sets filtered from various perspectives to train the dialogue model. 
However, without the view-wise guidance one by one, the union can not enforce the model to learn features biased towards different attributes effectively. Moreover, the union-based training will degrade to the traditional training when there are too many perspectives to consider. 

One can also use sequential learning \citep{DBLP:journals/corr/abs-1811-01088} or ensemble learning \citep{DBLP:journals/widm/SagiR18} to fit all filtered training sets gradually or parallelly, respectively. 
Unfortunately, the former suffers from catastrophic forgetting, i.e., knowledge learned from old training sets is always damaged by new training sets, and the latter will lead to serious knowledge interference \citep{DBLP:conf/eacl/PfeifferKRCG21}.

To avoid the problems of Data Filtering and meet the requirement of the dialogue system, in this work, we propose a \textit{multi-view attribute-enhanced} dialogue learning framework (\textbf{MAE}) to improve the conversational ability of the model from multiple perspectives effectively. 
Unlike the data filtering strategy that ignores the learning of non-target dialogue attributes, our framework aims at enhancing the target attribute without weakening any other attribute. 
It consists of one base model and multiple adapters. The base model is first pre-trained on the raw training set, allowing the framework to learn various features roughly. Then each adapter \citep{DBLP:conf/icml/HoulsbyGJMLGAG19} is fine-tuned on the sub-set selected by the corresponding scoring method, which enables the framework to further capture more features related to the target attributes without erasing any feature learned earlier. 
In order to generate the responses regarded as high-quality from multiple perspectives, we design two mechanisms to integrate complementary features in different adapters. 
The first one, \textit{Adaptive Fusion} (\textbf{AF}), ensembles multi-view features through the weighted average in inference after all adapters are fine-tuned in parallel, which keeps the adapters independent and plug-and-play. 
However, due to the knowledge interference among adapters, the features learned by one adapter may damage the features from other adapters, resulting in a sub-optimal integration. 
The second one, \textit{Progressive Fusion} (\textbf{PF}), constructs an incremental integration process through knowledge distillation \citep{DBLP:journals/corr/HintonVD15}, which enforces each new adapter to learn features complementary to those learned by previous adapters. 
Besides, the capacity of the framework will not increase significantly, as each adapter consists of very few parameters.

Our contributions are summarized as follows: 
(1) We propose a robust attribute-enhanced dialogue learning framework that strengthens the attribute-related features effectively while avoiding the problems of data filtering. 
(2) To improve the response quality more comprehensively, we further design two fusion mechanisms, AF and PF, to combine multi-view features from different adapters in inference and training, respectively. 
(3) We conduct extensive experiments to verify the effectiveness of MAE and provide a detailed analysis of feature learning and fusion. 

\section{Background}
\label{backgd}

\subsection{Dialogue Generation Models} 

Previous work enhancing the quality of responses falls into three major categories. 
The first redesigns the model structure to facilitate the modeling of the dialogue pairs \citep{DBLP:conf/aaai/SerbanKTTZBC17,CMHAM-TaoChongyang-2018,SpaceFusion-GaoXiang-2019}. 
The second further proposes the objectives that aligns with the goals of the conversation more effectively, such as MMI \citep{MMI-LiJiwei-2016}, CVAE \citep{VHRED-Serban-2017,kgCVAE-ZhaoTiancheng-2017,DialogWAE-GuXiaodong-2019,DBLP:conf/acl/SunFLLL20}, RL \citep{RLdialoguesys-LiJiwei-2016,RL-Seq2seqCo-Zhang2018,RL-P2BOT-Liu2020}, and GAN \citep{GAN-GANAEL-Xu2017,DPGAN-XuJingjing-2018,DBLP:conf/aaai/FengCLY20}. 
The third tries to endow the responses with topic \citep{TopicAware-XingChen-2017,DBLP:conf/emnlp/FengRCSLS20}, emotion \cite{Emotional-Zhouhao-2018,DBLP:conf/acl/RashkinSLB19}, and persona \citep{Persona-QianQiao-2017,PersonaChat-facebook-2018,DBLP:conf/acl/SongWZLL20}.
Recently, another line of work \citep{DialoGPT-Zhang-2020,Blender-Roller-2020,Meena-Adiwardana-2020,PLATO2-Bao-2020}, called the pre-trained dialogue model, relies on an efficient neural network and large-scale datasets to further improve the response quality.

\subsection{Data Filtering for Dialogue Learning} 
Many studies \citep{DBLP:journals/corr/FanTQBL17,DBLP:conf/icml/RenZYU18,DistributionDialog-Baheti-2018} argue that the quality of samples has a significant impact on the model performance. 
Recently, some researchers has considered the sample quality into the dialogue learning by a data manipulation strategy called Data Filtering, which discards samples that are regarded as low-quality by a scoring method. \citet{FilterEntropy-Csaky-2019} introduces an entropy-based scoring method to remove generic utterances from the training data. \citet{FilterSpecificity-See-2019} designs a scoring method to measure the specificity of samples. \citet{FilterConsistency-Akama-2020} combines the cosine distance and the keyword co-occurrence of the dialogue pairs to evaluate the sample coherency jointly. \citet{DBLP:conf/cikm/ShenZSCZZ21} proposes a fusing approach for data filtering by linearly combining seven scoring methods via Bayesian Optimization \citep{DBLP:journals/corr/abs-1012-2599}.
Unlike Data Filtering, our work designs a novel data manipulation framework to enhance the target attributes without sacrificing the feature learning of other dialogue attributes. 
There is another line of work \citep{DBLP:conf/sigdial/LisonB17,DBLP:conf/ijcai/ShangFPFZY18,DBLP:conf/acl/CaiCSZZY20}, named Data Weighting, that assigns the training samples with different weights, which is out of the scope of this work. All of them compute the weighting scores by a trainable model, which can also be replaced by the above scoring methods.

\subsection{Adapters in NLP} 

As a light-weight module, the adapter can be embedded into each layer of the pre-trained model to learn task-specific knowledge more efficiently, such as language features \citep{DBLP:conf/icml/HoulsbyGJMLGAG19,DBLP:journals/corr/abs-2002-01808} and multilingual features \citep{DBLP:conf/emnlp/BapnaF19,DBLP:conf/emnlp/PhilipBGB20,DBLP:conf/emnlp/PfeifferVGR20,DBLP:conf/nips/GuoZXWCC20,DBLP:conf/acl/RustPVRG20}. Previous work aims to transfer knowledge in the pre-trained model for the downstream tasks while avoiding catastrophic forgetting. 
Different from that, we further explore injecting view-specific knowledge into adapters and fusing multi-view knowledge to enhance the pre-trained model. 
\citet{DBLP:conf/eacl/PfeifferKRCG21} also tries to integrate knowledge from adapters, but the fusion layer will bring too many parameters as the number of adapters increases. Our method does not require extra parameters apart from adapters for knowledge fusion and therefore enables flexible integration.

\section{Framework}
\label{method}

The proposed framework offers a novel data manipulation paradigm to enhance the conversational skills of the model robustly and comprehensively. In this section, we first describe the collection process of view-specific training sub-sets, and then elaborate on how the adapters capture attribute-related features based on the corresponding sub-sets. Finally, two fusion mechanisms are introduced to integrate multi-view features. 
Algorithm \ref{appendix algorithm}, provided in the Appendix, shows full training details.

\subsection{View-Specific Collection}

\begin{figure}[t]
\centering
\includegraphics[width=0.8\linewidth]{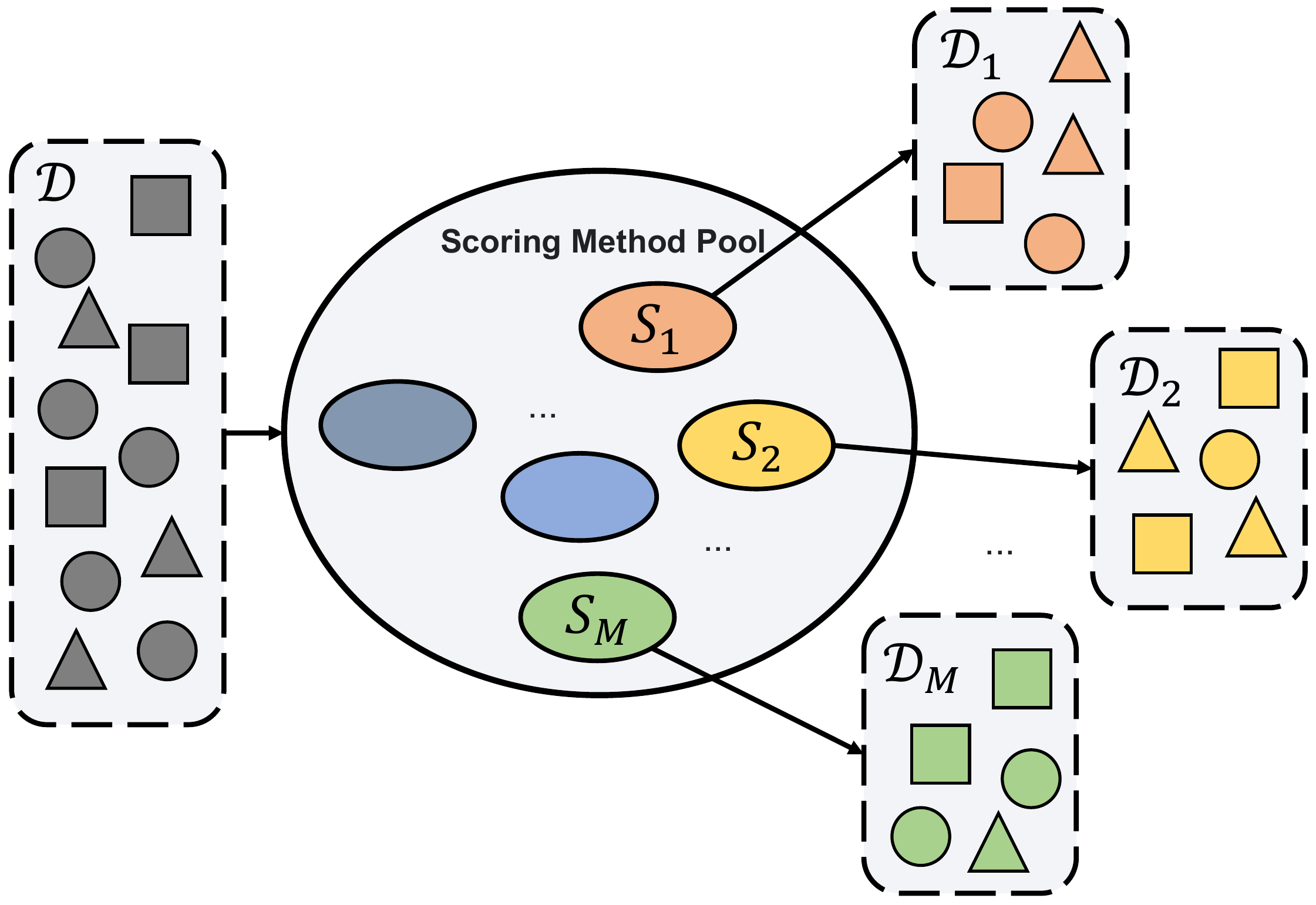}
\caption{The scoring method pool in MAE.}
\label{fig.md2f_filter}
\end{figure}

Previous filtering-based work aims at discarding the \textit{noisy} samples discriminated by the proposed scoring method. 
However, we have verified that the training sample with a high score given by any other scoring method can still be regarded as high-quality. Directly removing the \textit{noisy} samples will weaken the learning of features in other aspects. 
To tackle this problem, our framework measures the sample quality from multiple perspectives. 
Concretely, we construct a pool of scoring methods ($\mathcal{S}_{1}, \mathcal{S}_{2}, \cdots, \mathcal{S}_{M}$), shown in Figure~\ref{fig.md2f_filter}. After flowing through this pool, the raw training samples $\mathcal{D}$ are reorganized into multiple view-specific training sub-sets ($\mathcal{D}_{1}, \mathcal{D}_{2}, \cdots, \mathcal{D}_{M}$) based on a certain selection proportion. Note that each sample can be assigned to multiple sub-sets as it may obtain high scores from more than one scoring method. 
Differing from previous work that only treats the selected samples as high-quality data, we argue that different sub-sets $\mathcal{D}_{m}$ provide view-wise guidance for the learning of attribute-related features. Intuitively, the data selection can be regarded as an implicit cluster based on the dialogue attributes.

\begin{figure}[t]
\centering
\includegraphics[width=0.8\linewidth]{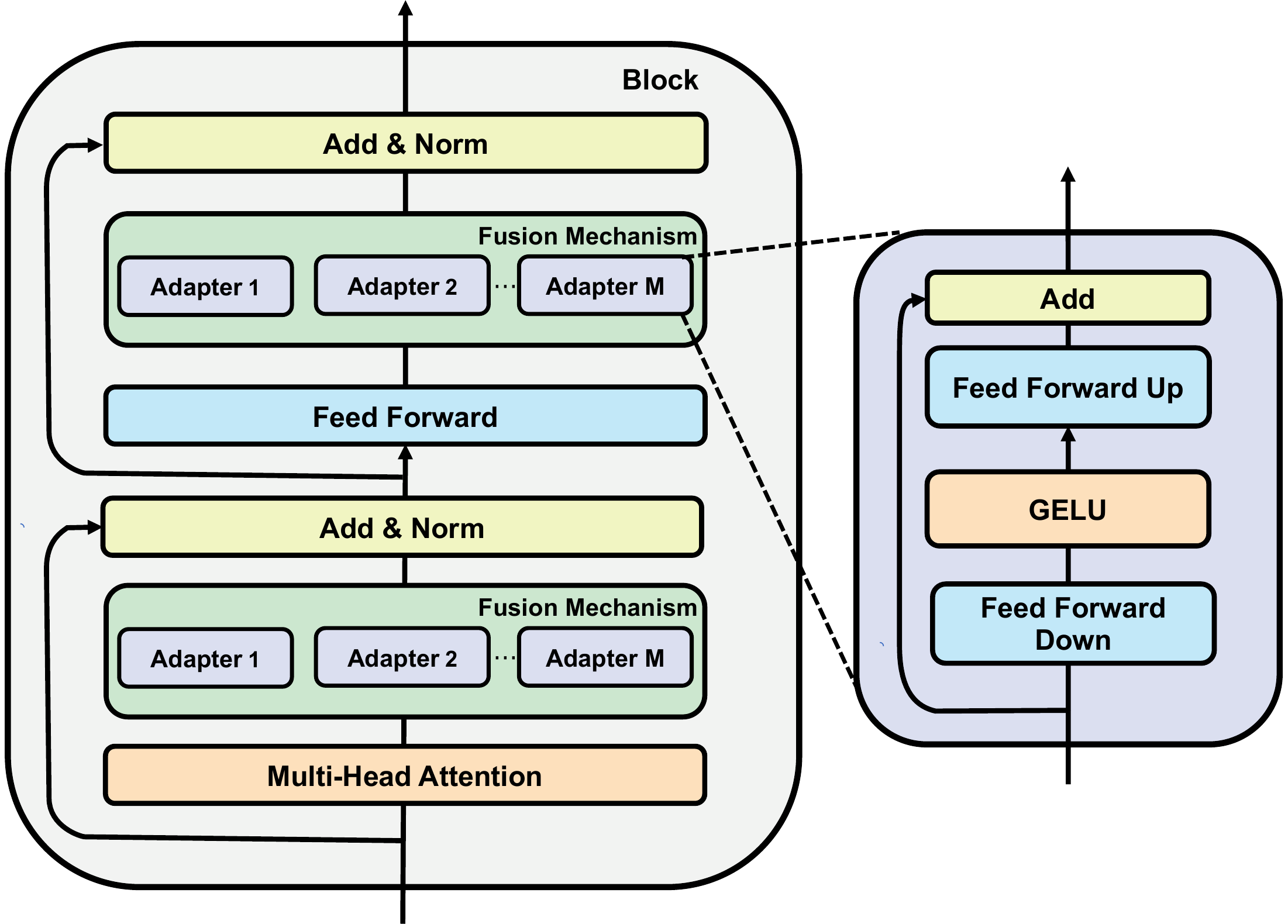}

\caption{The model architecture of MAE (Left) and the adapter layer (Right).}
\label{fig.md2f_framework}
\end{figure}

\subsection{View-Specific Dialogue Learning}

Previous work directly trains the model with the filtered training sets $\mathcal{D}_{m}$, which leads to information loss in other views, i.e., ignores the feature learning of non-target dialogue attributes. 
To address these issues, we use a two-stage training strategy that first warms up the base model on the raw training set $\mathcal{D}$ and then introduces the adapters to capture attribute-related features from all view-specific training sub-sets $\mathcal{D}_{m}$.

\subsubsection{Pre-Training Base Model}

We employ the standard Transformer architecture \citep{Transformer-Vaswani-2017} without adapter layers as the base model . The goal of the first-stage training is to enable the base model to access all training samples and learn the basic features. 
Formally, we maximize the probability $P_{\theta}(r|q)$ of each training sample ($q$,$r$) by optimizing the negative log likelihood (NLL) defined as:
\begin{equation}
    \label{eq:loss}
    \mathcal{L}_{nll}(\theta)=-\sum_{i=1}^{|r|} \log P_{\theta}\left(t^{r}_{i} \mid t^{r}_{<i}, q\right),
\end{equation}
where $|r|$ is the length of $r$, and $\theta$ represents the parameters of the base model.  
\subsubsection{Fine-Tuning Adapters} 
\label{fine-tuning}

In the second stage, we introduce the light-weight adapter layers into both encoder and decoder of the base model. 
Following \citet{DBLP:conf/icml/HoulsbyGJMLGAG19}, each encoder block of the base model contains two adapter layers (three in decoder), and each adapter layer consists of one bottleneck module (see Figure~\ref{fig.md2f_framework}). 
The attribute-related features $z^{m}$ can be represented as: 
\begin{equation}
    z^{m} = gelu(z \cdot w_{Down}^{m}) \cdot w_{Up}^{m}+z,
\end{equation}
where $z$ is the output of the previous layer, and $w_{Down}^{m}$ and $w_{Up}^{m}$ are the parameters of one adapter layer. 
The parameters of the base model are fixed, and the adapters parameters $\phi_{S_m}$ are independently fine-tuned on the corresponding view-specific training set $\mathcal{D}_{m}$ by $\mathcal{L}_{nll}(\phi_{m})$. 
Due to the light-weight structure, the increasing number of adapters will not bring excessive parameters. 
In addition, each adapter can learn attribute-related features without the distraction of the \textit{noisy} samples while avoiding catastrophic forgetting.

\subsection{Multi-View Attributes Fusion}

The goal of the dialogue system is to generate responses that can perform well on multiple dialogue attributes. 
In this section, we introduce two fusion mechanisms, Adaptive Fusion (AF) and Progressive Fusion (PF) to effectively exploit the multi-view knowledge for generating high-quality responses.

\subsubsection{Adaptive Fusion}
\begin{figure}[t]
\centering
\includegraphics[width=0.7\linewidth]{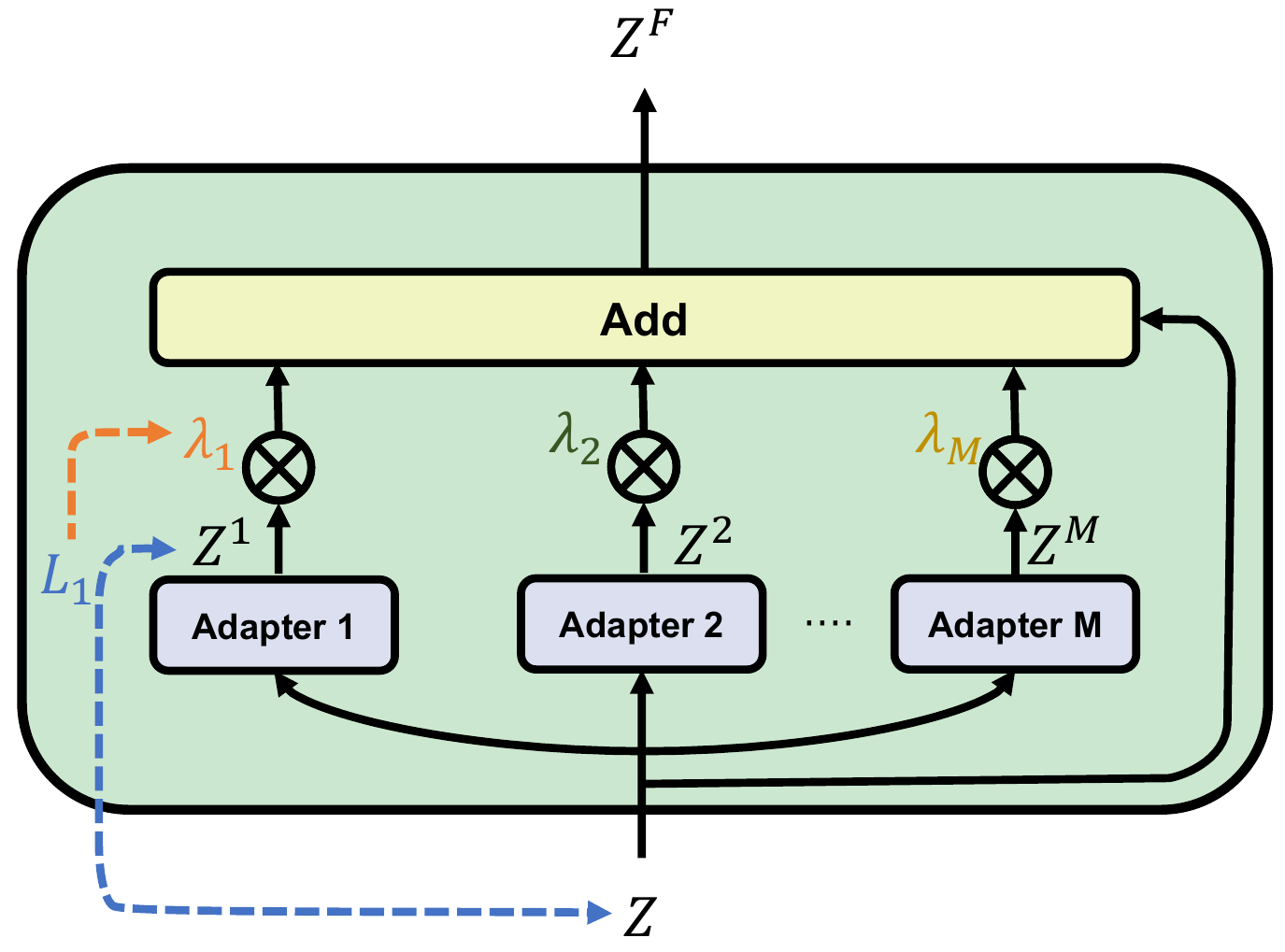}
\caption{Adaptive Fusion (AF) Mechanism}
\label{fig.md2f_af}
\end{figure}

For the adaptive fusion mechanism, all adapters can be fine-tuned in parallel. The fusion process, shown in Figure~\ref{fig.md2f_af}, combines multi-view features from different adapters in inference through a weighted average:
\begin{equation}
\label{eq:af_zandlambda}
    z^F = \sum_{m=1}^M\lambda_m z^{m}, \lambda_m = \frac{||z^{m}-z||_1}{\sum_{m}^M||z^{m}-z||_1}
\end{equation}
where $z^F$ is the output of the fusion mechanism, and $\lambda_m$ is the coefficient calculated by the L1-distance between $z^{m}$ and $z$. Inspired by \citet{DBLP:conf/icml/GuanWZCH019}, we take the distance between the input and output of view-specific adapter layer as the importance degree of the extracted features. See the Adaptive Weight Study for the analysis of its effectiveness.
The larger the distance is, the more the model needs these extracted features from the corresponding adapter layer for improving the overall quality of responses. We choose L1-distance rather than other types of distances due to its computational efficiency and higher discrimination. The AF can keep the adapters independent and plug-and-play, and conducts the layer-wise fusion that is more effective than ensemble learning. Yet, it may face the problem of knowledge interference that will affect the quality of generated responses.

\subsubsection{Progressive Fusion}
\begin{figure}[t]
\centering
\includegraphics[width=0.7\linewidth]{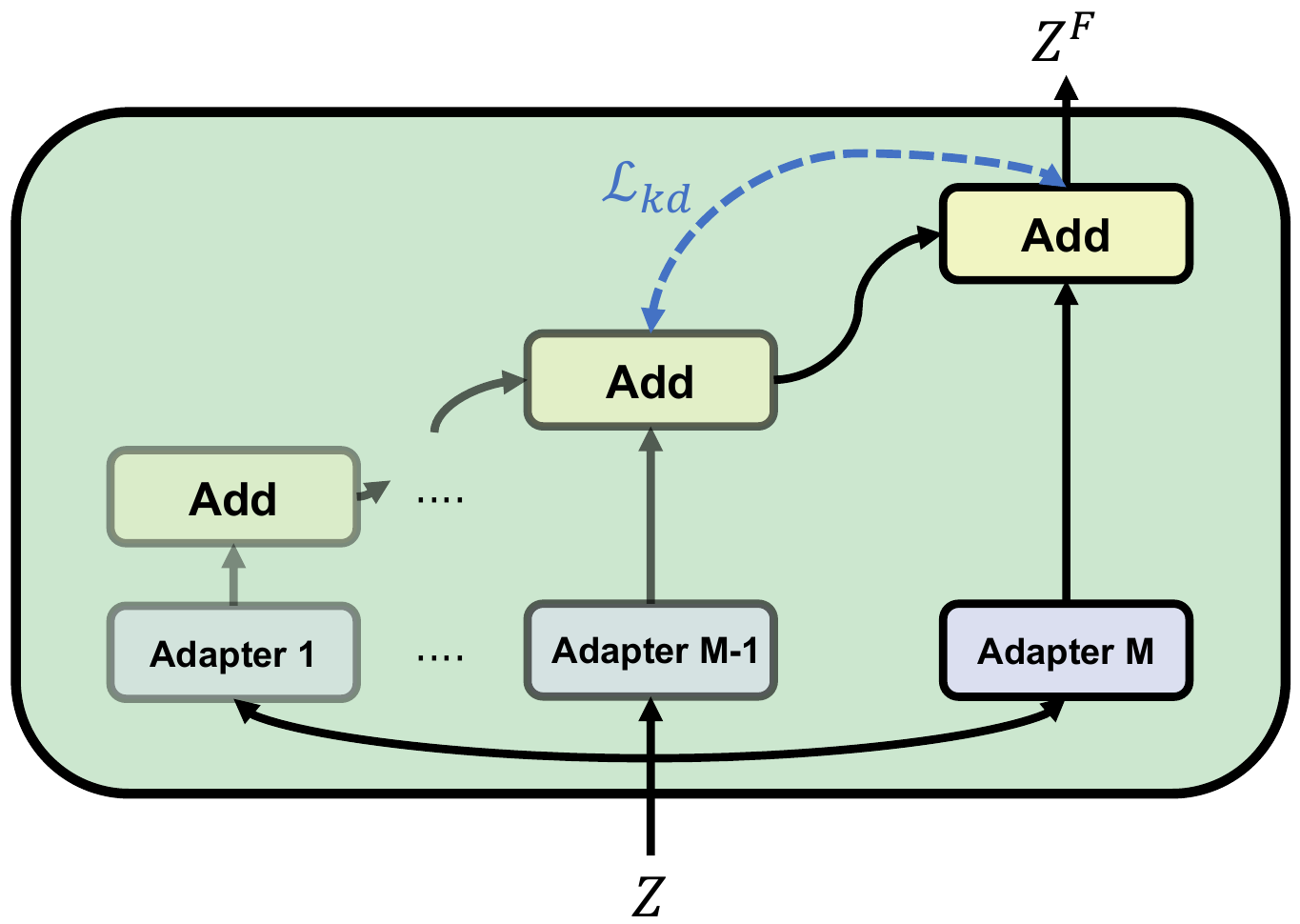}
\caption{Progressive Fusion (PF) Mechanism}
\label{fig.md2f_pf}
\end{figure}

The PF integrates multi-view features smoothly during training rather than inference, shown in Figure~\ref{fig.md2f_pf}, which requires the adapters to be fine-tuned sequentially. 
Each new adapter not only learns from the corresponding training set but also is enforced to find features complementary to those learned by previous adapters. 
Therefore, we use knowledge distillation to align the predictions of the base model with old adapters and the base model with both new and old adapters, which can be formulated as: 
\begin{align}
\nonumber \mathcal{L}_{kd}(\phi_n)&=- \sum_{i=1}^{|r|} \sum_{j=1}^{|\mathcal{V}|} P_{(\theta,\phi_{p})}\left(t_{i}^r=j \mid t_{<i}^r, q\right) \\
& \cdot \log P_{(\theta,\phi_p,\phi_n)}\left(t_{i}^r=j \mid t_{<i}^r, q\right), 
\end{align}
where $\phi_n$ is the parameters of the new adapter, $\phi_p$ is the frozen parameters of previous adapters, and $|V|$ denotes the vocabulary size. The final objective for training the new adapter is:
\begin{align}
    \label{eq:pf}
    \mathcal{L}(\phi_n)=\mathcal{L}_{nll}(\phi_n)+\lambda^{kd} \mathcal{L}_{kd}(\phi_n),
\end{align}
where $\lambda^{kd}$ represents the weight of $\mathcal{L}_{kd}$, $\lambda^{kd} = 1-\frac{current\_epoch}{total\_epoch}$, which decreases linearly during training. 
In this way, we give a strong constraint at the beginning of training to prevent the learned features from conflicting with features of previous adapters. 
And then, we reduce this constraint linearly to allow the new adapter to learn view-specific knowledge. 
Therefore, the knowledge of the adapters can be incorporated into the base model gradually while alleviating knowledge interference, but it will reduce the independence of adapters.

\begin{table*}[t]

    \centering
    \small
    \begin{tabular}{l|c c c c c|c c c c c}
    \toprule
        Models & Dist-1  & Dist-2  & KL-1 & KL-2 &BLEU & Dist-1  & Dist-2  & KL-1 & KL-2 &BLEU\\ \midrule
        Transformer & 0.0216 & 0.0728 & 1.67  & 1.66 & 0.292 &0.0157  & 0.0410 & 2.37  & 1.72  & 0.336 \\ \midrule
        Filtering-Con & 0.0225  & 0.0752  & 1.62  & 1.08 & 0.330 & 0.0189  & 0.0569 & 2.01  & 1.70 & 0.322 \\
        Filtering-Ent & 0.0166  & 0.0462  & 2.16  & 1.77  & 0.307 & 0.0156  & 0.0429 & 2.42  & 1.73  & 0.337 \\ 
        Filtering-Spe & 0.0132  & 0.0465  & 2.20  & 2.24  & 0.244 & 0.0150 &0.0439   & 1.93  & 1.82  & 0.311 \\
        \midrule
        Weighting-Con & 0.0050 & 0.0078 & 4.46 & 2.39 & 0.343 &0.0044 & 0.0082 & 4.47 & 3.38 & 0.248 \\
        Weighted-Ent & 0.0167 & 0.0455 & 2.02 & 1.54 & 0.341 &0.0185 & 0.0516 & 2.29 & \textbf{1.46} & \textbf{0.345} \\
        Weighted-Spe & 0.0156 & 0.0469 & 2.11 & 1.80 & 0.313 &0.0100 & 0.0265 & 3.01 & 2.46 & 0.291 \\
        \midrule
        MAE-AF & \underline{0.0434}  & \textbf{0.1522}  & \textbf{0.88}  & \underline{0.75} & \underline{0.383} & \underline{0.0204}  & \underline{0.0660} & \underline{1.83}  & 1.55 & 0.335 \\
        MAE-PF & \textbf{0.0463}  & \underline{0.1511} & \underline{0.94}  & \textbf{0.68}  & \textbf{0.392} & \textbf{0.0217}  & \textbf{0.0676} & \textbf{1.80}  & \textbf{1.46} & \underline{0.339}\\
    \bottomrule
    \end{tabular}
        \caption{Results of automatic evaluations on DailyDialog (Left) and OpenSubtitles (Right). The best/second-best results are \textbf{bold}/\underline{underlined}. For KL-\{1,2\}, lower is better.}

    \label{tb:main}
\end{table*}

\section{Experimental Setup}
\label{exp}

\begin{table*}[t]

    \centering
    \small
    \renewcommand\tabcolsep{2.5pt}
    \begin{tabular}{l|lll|lll|lll}
     \toprule
      \multirow{2}{*}{vs. Models} & \multicolumn{3}{c|}{Informativeness (\%)} & \multicolumn{3}{c|}{Relevance (\%)} & \multicolumn{3}{c}{Fluency (\%)} \\ 
      &  \makecell[c]{Win}  & \makecell[c]{Lose}  & \makecell[c]{Tie}  & \makecell[c]{Win}  & \makecell[c]{Lose}  & \makecell[c]{Tie}  & \makecell[c]{Win}  & \makecell[c]{Lose}  & \makecell[c]{Tie}   \\ \midrule
      Transformer &  36.0 / 33.3        &   15.3 / 14.0      &  48.7 / 52.7       &  38.0 / 36.0        &   \ \ 3.3 / \ \ 2.7      &   58.7 / 61.3      &  46.0 / 40.7       &   \ \ 5.3 / \ \ 4.0       &  48.7 / 55.3      \\
 Filtering-Con  &    61.3 / 59.3      &  \ \ 2.0 / \ \ 8.7       &  36.7 / 32.0       &44.7 / 40.0 &   \ \ 5.3 / \ \ 2.0        &  50.0 / 58.0       & 38.7 / 36.0        &  \ \ 8.7 / \ \ 4.7        &  53.3 / 59.3      \\
 Filtering-Ent & 56.0 / 52.0         & \ \ 6.7 / 12.7        & 37.3 / 35.3       & 52.0  / 47.3         & \ \ 3.3 / \ \ 2.0        & 44.7 / 50.7        & 32.7 / 30.0         & \ \ 7.3 / \ \ 6.0        & 60.0 / 64.0       \\
  Filtering-Spe &  26.7 / 30.7        & 19.3 / 24.7         & 54.0 / 44.7        &  45.3 / 48.0      & \ \ 6.0 / \ \ 1.3        &  48.7 / 50.7       & 47.3 / 40.0        & \ \ 2.7 / \ \ 2.7         & 50.0 / 57.3       \\
  Weighting-Ent &  52.0 / 44.0        & \ \ 6.7 / 12.0        & 41.3 / 44.0         & 45.3   / 45.3      & \ \ 6.7 / \ \ 4.0          &  48.0 / 50.7        &  22.7 / 25.3        & \ \ 8.0 / \ \ 6.7          &  69.3 / 68.0       \\
  
\midrule
 Transformer & 46.7 / 56.3 & \ \ 6.3 / \ \ 2.7 & 47.0 / 41.0 & 35.0 / 38.0 & 14.7 / \ \ 8.0 &     50.3 / 54.0 & 27.0 / 34.3 & 24.0 / 10.7 & 49.0 / 55.0 \\
 Filtering-Con &  29.0 / 35.0 & 24.7 / 13.0 & 46.3 /  52.0&    29.3 / 29.0      &  21.3 / 12.0       &    49.3 / 59.0     &  23.0 /  31.0     &  21.3 /  10.0      &  55.7 /  59.0    \\
 Filtering-Ent &   52.0 / 59.0     &17.3 / \ \ 3.0       &  30.7 / 38.0     &   42.3 / 42.3      &  13.7 / \ \ 4.0       &  44.0 / 53.3      &  36.0 / 47.0      &   14.3 / \ \ 5.7     &   49.7 / 47.3    \\
 Filtering-Spe &   41.0 / 51.3     &  19.0 / \ \ 9.7    &  40.0 / 40.0      &   36.7 / 42.3      &  \ \ 7.7 / \ \ 3.7        &  55.7 / 54.0  &   43.0 / 61.0   &   \ \ 7.3 / \ \ 3.0     &   49.7 /  36.0   \\
 Weighting-Ent &   62.7 / 68.0      &  11.0 / \ \ 4.3       &  26.3 / 27.7      &   42.3 / 43.3      &   11.3 / \ \ 7.6      &  46.3 / 49.0      &  30.0 / 44.7      &   18.3 / \ \ 8.7      &   51.7 / 46.7    \\
    \bottomrule
    \end{tabular}
        \caption{Results of human evaluations on DailyDialog (Top) and OpenSubtitles (Bottom). A/B in each table cell refer to the results of MAE-AF/MAE-PF, respectively. Our framework has a higher win rate than baselines.}
    \label{tb:human eval}
\end{table*}

\subsection{Datasets and Baselines}
\label{datasets and baselines}
We compare the proposed framework with one basic approach and three state-of-the-art filtering-based approaches on two open-domain dialogue datasets, DailyDialog \citep{dailydialog2017} and OpenSubtitles \citep{opensubtitles2009}. 

The basic approach trains the dialogue model on the entire training set. 
These three filtering-based approaches (\textbf{Filtering}) use the corresponding scoring methods that reflects the \textit{Consistency} (\textbf{Con}) \citep{FilterConsistency-Akama-2020}, \textit{Entropy\_Src} (\textbf{Ent}) \citep{FilterEntropy-Csaky-2019}, and \textit{Specificity} (\textbf{Spe}) \citep{FilterSpecificity-See-2019} of the dialogue pairs, respectively, to measure the quality of samples and discard the \textit{noisy} samples with low scores. Please refer to the Appendix~\ref{details for scoring methods} for the details of three high-quality automatic scoring methods. 
In addition, we also compare the \textbf{Weighting} approaches \citep{DBLP:conf/sigdial/LisonB17,DBLP:conf/ijcai/ShangFPFZY18,DBLP:conf/acl/CaiCSZZY20} although they can be seen as another line of data manipulation work. We replace their original weighting models with above three scoring methods to verify whether they are suitable for weighting approach.
Please refer to the Appendix~\ref{details for dataset} and \ref{details for scoring methods} for the details of datasets and scoring methods. 


\subsection{implementation details}
\label{implementation details}
Following previous work \citep{FilterEntropy-Csaky-2019,FilterConsistency-Akama-2020}, we take the Transformer-based dialogue model \citep{Transformer-Vaswani-2017} as the underling model for all approaches.

The settings of Transformer is consistent with \citet{FilterEntropy-Csaky-2019}, and please refer to the Appendix~\ref{training details} for the details. 

Our framework, including AF and PF, uses the same data \textit{selection} ratio of 50\% as \citet{FilterConsistency-Akama-2020} by three mentioned scoring methods. 
Note that the filtering-based baselines set the data \textit{filtering} ratio to 20\% due to their lower performance with 50\% ratio on the above datasets (see the Section~\ref{data volume} for the comparison). 
The fusion order of PF is chosen randomly (see the Appendix~\ref{appendix:order} for the analysis).

\subsection{Evaluation}
To comprehensively evaluate the quality of the generated responses, we conduct both automatic and human evaluations. 
The former employs three count-based metrics, \textbf{Dist-\{1,2\}}, \textbf{KL-\{1,2\}}, and \textbf{BLEU}, to reflect the linguistic quality, e.g., Dist and KL for the diversity and the distribution distance of n-grams, respectively. 
The latter focuses on more challenging semantic aspects, i.e., \textbf{Informativeness}, \textbf{Relevance}, and \textbf{Fluency}. 
Please refer to the Appendix~\ref{training details} for the details of the above metrics. 

\paragraph{Automatic evaluation.}

The automatic results in Table \ref{tb:main} show that our framework outperforms all baselines by a significant margin on both datasets, demonstrating the superiority of fusing multi-view features. 
MAE-PF obtains better results than MAE-AF on OpenSubtitles, which verifies that the PF mechanism can integrate multi-view features more smoothly. 
There is no noticeable gap between MAE-AF and MAE-PF on DailyDialog. Because all samples in DailyDialog are human-written and high-quality in multiple perspectives \citep{dailydialog2017}, knowledge interference among different adapters is weak. 
In addition, Filtering-Con achieves better performance than other filtering baselines, consistent with the results in \citet{FilterConsistency-Akama-2020}. Compared with Transformer, all filtering baselines gain more improvements on OpenSubtitles than on DailyDialog. These phenomena indicate that the filtering-based approaches are sensitive to both the scoring methods and the overall quality of datasets. 
As for the weighting approach, we find that the performance of the three scoring methods has changed a lot: \textit{Entropy\_Src} gets the better result than \textit{Consistency}, which shows that this kind of approaches has a heavy dependence on the scoring methods.

In contrast, our framework can avoid these problems due to the novel attribute-enhanced mechanism that does not damage the feature learning of the non-target dialogue attributes.

\paragraph{Human Evaluation} 

For each dataset, we randomly select 100 samples from the test set, and three well-educated annotators are hired to judge which of the responses generated by MAE-AF/PF and baselines is better (i.e., win, lose or tie) in terms of above three metrics. For weighting approach, we only select \textit{Entropy\_src} as the scoring method due to its much higher automatic performance than the other two.
The results, shown in Table \ref{tb:human eval}, demonstrate that our framework obtains a higher win rate than baselines in terms of three semantic metrics on both datasets. Besides, Filtering-Con performs better than other filtering baselines in Relevance, implying that the data filtering is beneficial for the model to enhance the dialogue attributes related to the scoring method. 
We use Fleiss's kappa \citep{fleisskappa/measuring} to assess the inter-annotator agreement, and the results are 0.541 and 0.619 on DailyDialog and OpenSubtitles, respectively.

\section{Further Analyses}
\label{experimental results}

In this section, we further investigate the advantages of MAE by providing detailed analyses.
Unless otherwise stated, the analysis results are based on DailyDialog.

\begin{table}
    \centering
    \small
    \renewcommand\tabcolsep{3.0pt}
    \begin{tabular}{l c c c c c }
    \toprule
        Models & Dist-1  & Dist-2  & KL-1 & KL-2 &BLEU \\ \midrule
        Filtering-Con (80$\%$) & 0.0225  & 0.0752 & 1.62  & 1.08 & 0.330  \\
        Filtering-Con (50$\%$) & 0.0088  & 0.0306 & 2.74  & 2.01 & 0.314 \\
        Filtering-Ent (80$\%$) & 0.0166  & 0.0462 & 2.16  & 1.77 & 0.307 \\ 
        Filtering-Ent (50$\%$) & 0.0096  & 0.0255 & 2.63  & 2.17 & 0.315 \\ 
        Filtering-Spe (80$\%$) & 0.0132  & 0.0465 & 2.20  & 2.24  & 0.244 \\
        Filtering-Spe (50$\%$) & 0.0061  & 0.0210 & 3.62  & 3.17  & 0.199 \\ 
        MAE-AF (80$\%$)& 0.0383  & 0.1370  & 0.97 & 0.85 & 0.375  \\
        MAE-AF (50$\%$)& 0.0434  & 0.1522  & 0.88  & 0.75 & 0.383 \\
        MAE-PF (80$\%$) & 0.0403 & 0.1361 & 0.94  & 0.68  & 0.382 \\
        MAE-PF (50$\%$) & 0.0463 & 0.1511 & 0.97  & 0.78  & 0.392 \\
    \bottomrule
    \end{tabular}
    \caption{Impact of the selection ratio on the model performance.}

    \label{tb:volume}
\end{table}

\subsection{Ablation Study} 
\label{data volume} 

To analyze the effect of the selection ratio on the model performance, we train the filtering baselines and our framework with two different ratios on both two datasets.
From the results in Table \ref{tb:volume} on single dataset, we find that the filtering baselines achieve better performance with a ratio of 80\% than with a ratio of 50\%, and our framework still perform well with a ratio of 50\%. 
This phenomenon illustrates that previous filtering-based approaches have a risk of dropping too many samples that are regarded as high-quality and benefits the model learning. 
In addition, too high selection ratio will cause too much overlap between the view-specific sub-sets, which is not beneficial for adapters to learn features biased towards different attributes effectively.

\begin{figure}[t]
    \centering
    \includegraphics[scale=0.36]{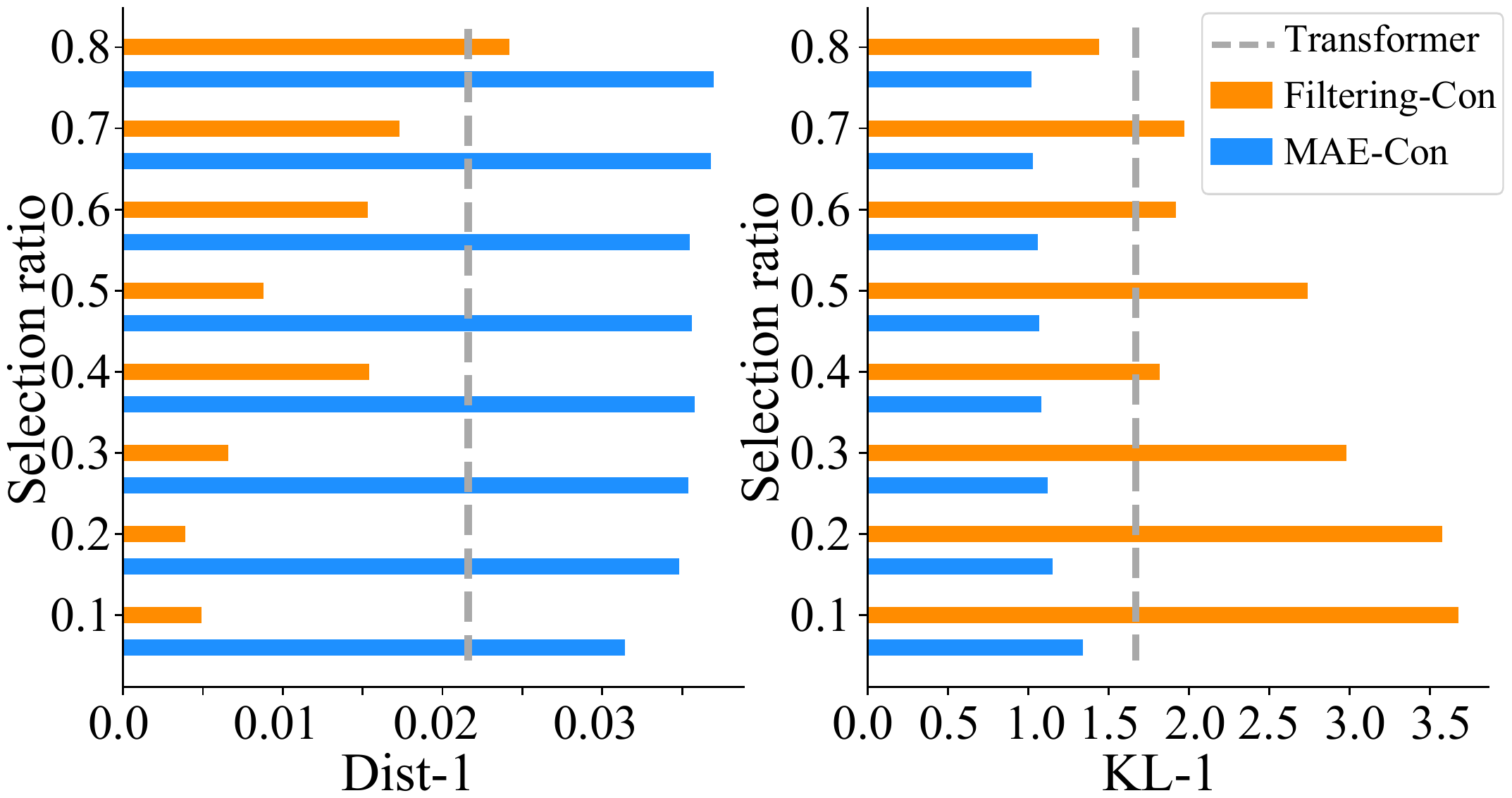}
    \caption{The variation of the model performance with respect to different selection ratios. MAE-Con consists of a pre-trained base model and an adapter fine-tuned on the sub-set that is selected from the \textit{Consistency} perspective.}
    \label{fig:a4}
\end{figure}

We also gradually decrease the selection ratio from 80\% to 10\% to observe the variation of the model performance. 
The results in Figure \ref{fig:a4} show that the performance of Filtering-Con is very unstable, even worse than Transformer, as the ratio varies. 
In contrast, MAE-Con has no significant changes in Dist-1 and KL-1, verifying the robustness of our attribute-enhanced mechanism. 
See Appendix~\ref{appendix:volume all} for the variation of other metrics. 

\subsection{View-specific Study} 
To verify that adapters can learn the corresponding attribute-related features from the view-specific sub-sets, we compare the generated response of MAE-Con with those of MAE-Spe in terms of \textit{Consistency} and \textit{Specificity}. 
Coherence (COH) \citep{FilterCoherence-Xu-2018} and Word Entropy (H-\{1,2\}) \citep{FilterEntropy-Csaky-2019} are adopted to assess the consistency and specificity of responses, respectively.
Besides, we also conduct human evaluations.
See the Appendix for the details of these metrics.

In Table \ref{tb:a3}, we find that MAE-Con indeed performs better than MAE-Spe in \textit{Consistency} and vice versa. 
Moreover, the results in Table \ref{tb:compare single attribute} indicate that our framework is also more effective than previous filtering approaches on improving the overall quality of responses, contributing to the novel attribute-enhanced mechanism.

\begin{table}[t]

    \centering
    \small
    \renewcommand\tabcolsep{3.5pt}
    \begin{tabular}{l | c c c | c c}
    \toprule
        Models & COH & H-1 & H-2 & Consistency & Specificity \\ \midrule
        MAE-Con & \textbf{0.729}   & 6.84   & 7.80 & \textbf{60.0\%} & 14.7\% \\ 
        MAE-Spe & 0.717   & \textbf{7.25}   & \textbf{8.23} & 8.0\% & \textbf{60.7\%} \\ 
    \bottomrule
    \end{tabular}
    \caption{Results of automatic (Left) and human (Right; win rate) evaluations of MAE-Con and MAE-Spe.}
    \label{tb:a3}
\end{table}

\begin{table}[t]

    \centering
    \small
    \renewcommand\tabcolsep{4.5pt}
    \begin{tabular}{l l l l l l}
    \toprule
        Models & Dist-1  & Dist-2   & KL-1 & KL-2  &BLEU  \\ \midrule
        Filtering-Con & 0.0225 & 0.0752 & 1.62 & 1.08 & 0.330 \\
        MAE-Con & \textbf{0.0356} & \textbf{0.1239} & \textbf{1.07} & \textbf{0.97}& \textbf{0.362} \\
        \midrule
        Filtering-Spe & 0.0132 & 0.0465 & 2.20 & 2.24 & 0.244 \\
        MAE-Spe & \textbf{0.0333} & \textbf{0.1156} & \textbf{1.20} & \textbf{1.29} & \textbf{0.330} \\
    \bottomrule
    \end{tabular}
    \caption{Comparison of the conventional filtering-based approach and MAE with a single adapter.}
    \label{tb:compare single attribute}
\end{table}

\begin{table}[b]
    \centering
    \small
    \renewcommand\tabcolsep{3.0pt}
    \begin{tabular}{l l l l l l}
    \toprule
        Models & Dist-1  & Dist-2   & KL-1 & KL-2  &BLEU  \\ \midrule

        Intersection  & 0.0085 & 0.0258 & 3.15  & 2.18 & 0.322 \\ 
        Union         & 0.0244 & 0.0739 & 1.56 & 1.34 & 0.353  \\ 
        Ensemble      & 0.0115 & 0.0291 & 2.92 & 2.19 & 0.289  \\
        Sequential    & 0.0313 & 0.1031 & 1.23 & 0.86 & 0.375  \\
        \midrule
         MAE-AF & \underline{0.0434}  & \textbf{0.1522}  & \textbf{0.88}  & \textbf{0.75} & \underline{0.383} \\
        MAE-PF  & \textbf{0.0463} & \underline{0.1511} & \underline{0.97}  & \underline{0.78}  & \textbf{0.392} \\       
    \bottomrule
    \end{tabular}
    \caption{Comparison of MAE-AF/PF and the conventional filtering-based approach with four fusion strategies.}
    \label{tb:fusion}
\end{table}

\subsection{Fusion Mechanism Study} 
We compare the proposed framework with the conventional filtering-based approach equipped with a variety of fusion strategies. 
\textit{Intersection} and \textit{Union} use the intersection and union of different sub-sets to train the vanilla Transformer, respectively. \textit{Ensemble} directly combines the outputs of three baselines (i.e., Filtering-xxx). \textit{Sequential} trains Transformer on different sub-sets one by one. 
Note that for scores of Con and Spe, the higher the better, but for Ent, the opposite. Therefore, \citet{DBLP:conf/cikm/ShenZSCZZ21} is not suitable here because it is hard to combine these scoring methods linearly. 
As shown in Table \ref{tb:fusion}, \textit{Intersection} gets the worst results due to the model trained with too few samples. 
\textit{Union} achieves almost the same performance as Filtering-Con, indicating that the union of view-specific sub-sets can not induce the model to learn features biased towards different dialogue attributes effectively. 
\textit{Sequential} performs better than \textit{Ensemble}, even Filtering-Con, but it is still weaker than MAE-AF/PF. 
It is because \textit{Ensemble} easily suffers from knowledge interference due to only fusing the outputs of models, and \textit{Sequential} is inevitably limited by catastrophic forgetting. 
However, our framework can learn and save view-specific knowledge independently, and fuse them smoothly.

\subsection{Adaptive Weight Study} 

Can the L1-distance between the input and output of the adapter layer reflect the importance of the adapter layer?

In Table \ref{tb:distance}, we report the average L1-distance calculated based on all adapter layers of each module, and give the corresponding model performance when deleting all adapter layers of each block in inference. 
The results demonstrate that the adapter layer with a larger L1-distance usually conducts a stronger impact on the model performance. 
The adapter layers in the bottom and top blocks are more important than those in the middle blocks. 

\begin{table}
    \centering
    \small
    \renewcommand\tabcolsep{0.6pt}
    \begin{tabular}{c| c c c| c c c c c}
    \toprule

        Block & Self-attn & ED-attn & FFN & Dist-1  & Dist-2  & KL-1 & KL-2 & BLEU \\ \midrule
        1 & 375.4  & 224.9  & 1278.9 & .0428 & .1512 & .91 & .76  & .390 \\
        2 & 315.9  & 229.4  & 698.0& .0437  & .1509 & .92  & .69  & .389 \\
        3 & 365.6  & 227.8  & 355.2& .0447  & .1545 & .90  & .69  & .391 \\ 
        4 & 281.4  & 217.1  & 312.0& .0450  & .1529 & .92  & .69  & .389 \\ 
        5 & 230.9  & 218.6  & 300.9& .0448  & .1538 & .88  & .64  & .399 \\ 
        6 & 392.7  & 292.0  & 516.7& .0336  & .1081 & 1.24  & 1.16  & .348 \\ 
    \bottomrule
    \end{tabular}

    \caption{Left: The average L1-distance of all adapter layers of each module. Right: The corresponding performance of MAE-AF when deleting all adapter layers of each block in inference. Self-attn, ED-attn, and FFN represent the self-attention module, the encoder-decoder attention module, and the feed-forward module of the decoder, respectively.}
    \label{tb:distance}
\end{table}

\begin{table}[t]

\centering
\footnotesize

\begin{tabular}{@{}p{\linewidth}@{}}
\toprule

\textbf{Input:} That sounds really neat. Is there an extra charge for these swim stations?  \\
\midrule
\textbf{Transformer:} Yes, it is.  \\

\textbf{MAE-AF:} Yes, it is. But it's a very expensive place. \\
\midrule
\midrule
\textbf{Input:} I just happen to have a question for you guys. Why do the Chinese cook the vegetables? You see what I mean is that most vitamin are destroyed when heated. \\
\midrule
\textbf{Transformer:} I ’ m sorry, sir. We' re going to have to take a lot of money to get in. \\

\textbf{MAE-AF:} I think you should have a good idea. \\
\bottomrule
\end{tabular}
\caption{Examples of the responses generated by Transformer and MAE-AF.}
\label{tab:dis case}
\vspace{-1.5\baselineskip}
\end{table}

\begin{figure}[ht]
    \centering
    \includegraphics[scale=0.335]{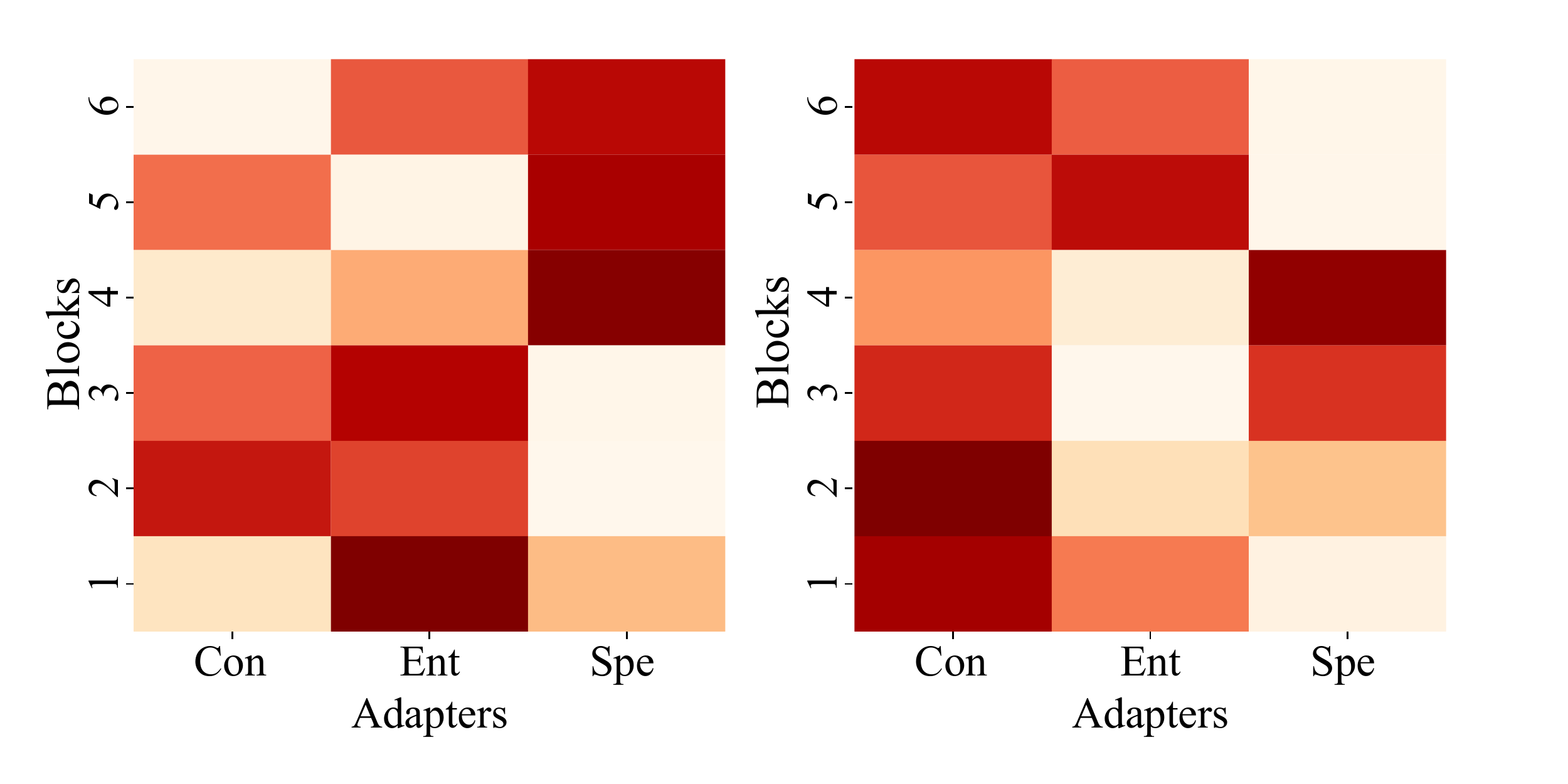}
    \caption{Heat map of the average weight of each type of adapter layers in each block. Left for the first example and Right for the second example in Table \ref{tab:dis case}.}
    \label{fig:heat}
\end{figure}

Which type of adapter layers is more important? 
We first provide two examples in Table~\ref{tab:dis case}. It can be seen that the response of MAE-AF is more informative and diverse than the response of Transformer in the first example, and more consistent with the context in the second example. 
Then, we visualize the weights of different types of adapter layers, shown in Figure \ref{fig:heat}. Surprisingly, the first heat map shows darker color in the adapter layers of Ent and Spe, and the second heat map shows darker color in the adapter layers of Con. 
This phenomenon verifies that the weights of adapters reflect the drawbacks of responses generated by Transformer, and our framework can adaptively adjusts the weights of adapters to enhance the related features to improve the overall quality of responses.

\subsection{Case Study}
Table \ref{tab:case study} presents some responses generated by the proposed framework and baselines. 
Transformer prefers generic and meaningless responses. The responses generated by filtering methods are usually corresponding to one perspective of dialogue. 
For instance, the Filtering-Ent approach often generates informative but irrelevant responses. 
In contrast, our AF and PF comprehensively consider the multiple perspectives, thus resulting in diverse and coherent responses. The results demonstrate the effectiveness of MAE.

\begin{table}[t]

\centering
\footnotesize
\begin{tabular}{@{}p{\linewidth}@{}}
\toprule

\textbf{Input:} Operator. Can I help you? \\
\midrule
\textbf{Transformer:} Yes, I'm calling about your company. \\
\textbf{Filtering-Con:} Yes, I'm here to see if it is going to be a job.  \\
\textbf{Filtering-Ent:} Yes, I'm going to be in a while. I'm going to have to see my company.  \\
\textbf{Filtering-Spe:} I’ m not sure, I’m not going to get a new one. \\
\midrule
\textbf{MAE-AF:} Yes, I am calling to tell you that I have been here for a week.  \\
\textbf{MAE-PF:} Yes, I am calling to tell you that I would like to help my company. \\
\midrule
\midrule
\textbf{Input:} I am keen on a coat on www ... com. But it requires having an online banking account. Is it necessary?  \\
\midrule
\textbf{Transformer:} Yes, it's a very good idea. \\
\textbf{Filtering-Con:} Yes, I do. I have a new job with you.  \\
\textbf{Filtering-Ent:} Yes, we have a very good job in the office. And we can take a new card and a new room.  \\
\textbf{Filtering-Spe:} I think you can get a good idea. \\
\midrule
\textbf{MAE-AF:} Yes, it is. But you have to pay a new one. It's a very good place to get to know some new York.  \\
\textbf{MAE-PF:} Yes, it is. But it's a very good place to go with a new York. \\

\bottomrule
\end{tabular}
\caption{Examples of generated responses.}
\label{tab:case study}

\end{table}

\section{Conclusion} 
In this work, we propose a novel multi-view attribute-enhanced dialogue learning framework that induces the model to enhance related knowledge along with the dialogue attributes and fuse them for the overall improvement of the response quality. 
We first collect various view-specific sub-sets from the raw training set. Then the adapters are introduced to learn and save more features biased towards different dialogue attributes. 
Finally, we design two fusion mechanisms, Adaptive Fusion and Progressive Fusion, to integrate multi-view knowledge of adapters in inference and training, respectively. The former makes the adapters plug-and-play, and the latter alleviates knowledge interference. 
The experimental results and analysis demonstrate that our framework learns attribute-related features, improves the model robustness due to the novel enhance mechanism, and integrates features of adapters effectively. 
Compared with previous data filtering approaches, it offers a new perspective to incorporate the sample quality into the model learning.

\bibliography{custom}

\begin{thebibliography}{66}
\expandafter\ifx\csname natexlab\endcsname\relax\def\natexlab#1{#1}\fi

\bibitem[{Adiwardana et~al.(2020)Adiwardana, Luong, So, Hall, Fiedel,
  Thoppilan, Yang, Kulshreshtha, Nemade, Lu, and Le}]{Meena-Adiwardana-2020}
Daniel Adiwardana, Minh{-}Thang Luong, David~R. So, Jamie Hall, Noah Fiedel,
  Romal Thoppilan, Zi~Yang, Apoorv Kulshreshtha, Gaurav Nemade, Yifeng Lu, and
  Quoc~V. Le. 2020.
\newblock \href {http://arxiv.org/abs/2001.09977} {Towards a human-like
  open-domain chatbot}.
\newblock \emph{CoRR}, abs/2001.09977.

\bibitem[{Akama et~al.(2020)Akama, Yokoi, Suzuki, and
  Inui}]{FilterConsistency-Akama-2020}
Reina Akama, Sho Yokoi, Jun Suzuki, and Kentaro Inui. 2020.
\newblock \href {https://doi.org/10.18653/v1/2020.emnlp-main.68} {Filtering
  noisy dialogue corpora by connectivity and content relatedness}.
\newblock In \emph{{EMNLP}}, pages 941--958.

\bibitem[{Baheti et~al.(2018)Baheti, Ritter, Li, and
  Dolan}]{DistributionDialog-Baheti-2018}
Ashutosh Baheti, Alan Ritter, Jiwei Li, and Bill Dolan. 2018.
\newblock \href {https://doi.org/10.18653/v1/d18-1431} {Generating more
  interesting responses in neural conversation models with distributional
  constraints}.
\newblock In \emph{{EMNLP}}, pages 3970--3980.

\bibitem[{Bao et~al.(2020)Bao, He, Wang, Wu, Wang, Wu, Guo, Liu, and
  Xu}]{PLATO2-Bao-2020}
Siqi Bao, Huang He, Fan Wang, Hua Wu, Haifeng Wang, Wenquan Wu, Zhen Guo,
  Zhibin Liu, and Xinchao Xu. 2020.
\newblock \href {http://arxiv.org/abs/2006.16779} {{PLATO-2:} towards building
  an open-domain chatbot via curriculum learning}.
\newblock \emph{CoRR}, abs/2006.16779.

\bibitem[{Bapna and Firat(2019)}]{DBLP:conf/emnlp/BapnaF19}
Ankur Bapna and Orhan Firat. 2019.
\newblock \href {https://doi.org/10.18653/v1/D19-1165} {Simple, scalable
  adaptation for neural machine translation}.
\newblock In \emph{{EMNLP-IJCNLP}}, pages 1538--1548.

\bibitem[{Bouma(2009)}]{nPMI-Bouma-2009}
Gerlof Bouma. 2009.
\newblock \href
  {https://svn.spraakdata.gu.se/repos/gerlof/pub/www/Docs/npmi-pfd.pdf}
  {Normalized (pointwise) mutual information in collocation extraction}.
\newblock In \emph{(GSCL)}, page 31–40.

\bibitem[{Bowman et~al.(2016)Bowman, Vilnis, Vinyals, Dai, J{\'{o}}zefowicz,
  and Bengio}]{VaeTextGeneration-Bowman-2016}
Samuel~R. Bowman, Luke Vilnis, Oriol Vinyals, Andrew~M. Dai, Rafal
  J{\'{o}}zefowicz, and Samy Bengio. 2016.
\newblock \href {https://doi.org/10.18653/v1/k16-1002} {Generating sentences
  from a continuous space}.
\newblock In \emph{CoNLL}, pages 10--21.

\bibitem[{Brochu et~al.(2010)Brochu, Cora, and
  de~Freitas}]{DBLP:journals/corr/abs-1012-2599}
Eric Brochu, Vlad~M. Cora, and Nando de~Freitas. 2010.
\newblock \href {http://arxiv.org/abs/1012.2599} {A tutorial on bayesian
  optimization of expensive cost functions, with application to active user
  modeling and hierarchical reinforcement learning}.
\newblock \emph{CoRR}, abs/1012.2599.

\bibitem[{Cai et~al.(2020)Cai, Chen, Song, Zhang, Zhao, and
  Yin}]{DBLP:conf/acl/CaiCSZZY20}
Hengyi Cai, Hongshen Chen, Yonghao Song, Cheng Zhang, Xiaofang Zhao, and Dawei
  Yin. 2020.
\newblock \href {https://doi.org/10.18653/v1/2020.acl-main.564} {Data
  manipulation: Towards effective instance learning for neural dialogue
  generation via learning to augment and reweight}.
\newblock In \emph{{ACL}}, pages 6334--6343. Association for Computational
  Linguistics.

\bibitem[{Chen and Cherry(2014)}]{DBLP:conf/wmt/ChenC14}
Boxing Chen and Colin Cherry. 2014.
\newblock \href {https://doi.org/10.3115/v1/w14-3346} {A systematic comparison
  of smoothing techniques for sentence-level {BLEU}}.
\newblock In \emph{Ninth Workshop on Statistical Machine Translation}, pages
  362--367.

\bibitem[{Chen et~al.(2017)Chen, Liu, Yin, and
  Tang}]{DBLP:journals/sigkdd/ChenLYT17}
Hongshen Chen, Xiaorui Liu, Dawei Yin, and Jiliang Tang. 2017.
\newblock \href {https://doi.org/10.1145/3166054.3166058} {A survey on dialogue
  systems: Recent advances and new frontiers}.
\newblock \emph{{SIGKDD} Explor.}, 19(2):25--35.

\bibitem[{Chen et~al.(2018)Chen, Ren, Tang, Zhao, and
  Yin}]{HVaeMN-ChenHongshen-2018}
Hongshen Chen, Zhaochun Ren, Jiliang Tang, Yihong~Eric Zhao, and Dawei Yin.
  2018.
\newblock \href {https://doi.org/10.1145/3178876.3186077} {Hierarchical
  variational memory network for dialogue generation}.
\newblock In \emph{{WWW}}, pages 1653--1662.

\bibitem[{Csaky et~al.(2019)Csaky, Purgai, and
  Recski}]{FilterEntropy-Csaky-2019}
Richard Csaky, Patrik Purgai, and G{\'{a}}bor Recski. 2019.
\newblock \href {https://doi.org/10.18653/v1/p19-1567} {Improving neural
  conversational models with entropy-based data filtering}.
\newblock In \emph{{ACL} {(1)}}, pages 5650--5669.

\bibitem[{Fan et~al.(2017)Fan, Tian, Qin, Bian, and
  Liu}]{DBLP:journals/corr/FanTQBL17}
Yang Fan, Fei Tian, Tao Qin, Jiang Bian, and Tie{-}Yan Liu. 2017.
\newblock \href {http://arxiv.org/abs/1702.08635} {Learning what data to
  learn}.
\newblock \emph{CoRR}, abs/1702.08635.

\bibitem[{Feng et~al.(2020{\natexlab{a}})Feng, Chen, Li, and
  Yin}]{DBLP:conf/aaai/FengCLY20}
Shaoxiong Feng, Hongshen Chen, Kan Li, and Dawei Yin. 2020{\natexlab{a}}.
\newblock \href {https://ojs.aaai.org/index.php/AAAI/article/view/6273}
  {Posterior-gan: Towards informative and coherent response generation with
  posterior generative adversarial network}.
\newblock In \emph{{AAAI}}, pages 7708--7715.

\bibitem[{Feng et~al.(2020{\natexlab{b}})Feng, Ren, Chen, Sun, Li, and
  Sun}]{DBLP:conf/emnlp/FengRCSLS20}
Shaoxiong Feng, Xuancheng Ren, Hongshen Chen, Bin Sun, Kan Li, and Xu~Sun.
  2020{\natexlab{b}}.
\newblock \href {https://doi.org/10.18653/v1/2020.emnlp-main.534} {Regularizing
  dialogue generation by imitating implicit scenarios}.
\newblock In \emph{{EMNLP}}, pages 6592--6604.

\bibitem[{Fleiss(1971)}]{fleisskappa/measuring}
Joseph~L Fleiss. 1971.
\newblock Measuring nominal scale agreement among many raters.
\newblock \emph{Psychological bulletin}, 76(5):378.

\bibitem[{Gao et~al.(2019)Gao, Lee, Zhang, Brockett, Galley, Gao, and
  Dolan}]{SpaceFusion-GaoXiang-2019}
Xiang Gao, Sungjin Lee, Yizhe Zhang, Chris Brockett, Michel Galley, Jianfeng
  Gao, and Bill Dolan. 2019.
\newblock \href {https://doi.org/10.18653/v1/n19-1125} {Jointly optimizing
  diversity and relevance in neural response generation}.
\newblock In \emph{{NAACL-HLT} {(1)}}, pages 1229--1238.

\bibitem[{Ghazvininejad et~al.(2018)Ghazvininejad, Brockett, Chang, Dolan, Gao,
  Yih, and Galley}]{FactKnowledge-Ghazvininejad-2018}
Marjan Ghazvininejad, Chris Brockett, Ming{-}Wei Chang, Bill Dolan, Jianfeng
  Gao, Wen{-}tau Yih, and Michel Galley. 2018.
\newblock \href
  {https://www.aaai.org/ocs/index.php/AAAI/AAAI18/paper/view/16710} {A
  knowledge-grounded neural conversation model}.
\newblock In \emph{{AAAI}}, pages 5110--5117.

\bibitem[{Gu et~al.(2019)Gu, Cho, Ha, and Kim}]{DialogWAE-GuXiaodong-2019}
Xiaodong Gu, Kyunghyun Cho, Jung{-}Woo Ha, and Sunghun Kim. 2019.
\newblock \href {https://openreview.net/forum?id=BkgBvsC9FQ} {Dialogwae:
  Multimodal response generation with conditional wasserstein auto-encoder}.
\newblock In \emph{{ICLR} (Poster)}.

\bibitem[{Guan et~al.(2019)Guan, Wang, Zhang, Chen, He, and
  Xie}]{DBLP:conf/icml/GuanWZCH019}
Chaoyu Guan, Xiting Wang, Quanshi Zhang, Runjin Chen, Di~He, and Xing Xie.
  2019.
\newblock \href {http://proceedings.mlr.press/v97/guan19a.html} {Towards a deep
  and unified understanding of deep neural models in {NLP}}.
\newblock In \emph{{ICML}}, volume~97, pages 2454--2463.

\bibitem[{Guo et~al.(2020)Guo, Zhang, Xu, Wei, Chen, and
  Chen}]{DBLP:conf/nips/GuoZXWCC20}
Junliang Guo, Zhirui Zhang, Linli Xu, Hao{-}Ran Wei, Boxing Chen, and Enhong
  Chen. 2020.
\newblock \href
  {https://proceedings.neurips.cc/paper/2020/hash/7a6a74cbe87bc60030a4bd041dd47b78-Abstract.html}
  {Incorporating {BERT} into parallel sequence decoding with adapters}.
\newblock In \emph{{NeurIPS}}.

\bibitem[{Hinton et~al.(2015)Hinton, Vinyals, and
  Dean}]{DBLP:journals/corr/HintonVD15}
Geoffrey~E. Hinton, Oriol Vinyals, and Jeffrey Dean. 2015.
\newblock \href {http://arxiv.org/abs/1503.02531} {Distilling the knowledge in
  a neural network}.
\newblock \emph{CoRR}, abs/1503.02531.

\bibitem[{Houlsby et~al.(2019)Houlsby, Giurgiu, Jastrzebski, Morrone,
  de~Laroussilhe, Gesmundo, Attariyan, and
  Gelly}]{DBLP:conf/icml/HoulsbyGJMLGAG19}
Neil Houlsby, Andrei Giurgiu, Stanislaw Jastrzebski, Bruna Morrone, Quentin
  de~Laroussilhe, Andrea Gesmundo, Mona Attariyan, and Sylvain Gelly. 2019.
\newblock \href {http://proceedings.mlr.press/v97/houlsby19a.html}
  {Parameter-efficient transfer learning for {NLP}}.
\newblock In \emph{{ICML}}, volume~97 of \emph{Proceedings of Machine Learning
  Research}, pages 2790--2799.

\bibitem[{Kingma and Ba(2015)}]{DBLP:journals/corr/KingmaB14}
Diederik~P. Kingma and Jimmy Ba. 2015.
\newblock \href {http://arxiv.org/abs/1412.6980} {Adam: {A} method for
  stochastic optimization}.
\newblock In \emph{3rd International Conference on Learning Representations,
  {ICLR} 2015, San Diego, CA, USA, May 7-9, 2015, Conference Track
  Proceedings}.

\bibitem[{Li et~al.(2016{\natexlab{a}})Li, Galley, Brockett, Gao, and
  Dolan}]{MMI-LiJiwei-2016}
Jiwei Li, Michel Galley, Chris Brockett, Jianfeng Gao, and Bill Dolan.
  2016{\natexlab{a}}.
\newblock \href {https://doi.org/10.18653/v1/n16-1014} {A diversity-promoting
  objective function for neural conversation models}.
\newblock In \emph{{HLT-NAACL}}, pages 110--119.

\bibitem[{Li et~al.(2016{\natexlab{b}})Li, Monroe, Ritter, Jurafsky, Galley,
  and Gao}]{RLdialoguesys-LiJiwei-2016}
Jiwei Li, Will Monroe, Alan Ritter, Dan Jurafsky, Michel Galley, and Jianfeng
  Gao. 2016{\natexlab{b}}.
\newblock \href {https://doi.org/10.18653/v1/d16-1127} {Deep reinforcement
  learning for dialogue generation}.
\newblock In \emph{{EMNLP}}, pages 1192--1202.

\bibitem[{Li et~al.(2017)Li, Su, Shen, Li, Cao, and Niu}]{dailydialog2017}
Yanran Li, Hui Su, Xiaoyu Shen, Wenjie Li, Ziqiang Cao, and Shuzi Niu. 2017.
\newblock \href {https://www.aclweb.org/anthology/I17-1099/} {Dailydialog: {A}
  manually labelled multi-turn dialogue dataset}.
\newblock In \emph{{IJCNLP(1)}}, pages 986--995.

\bibitem[{Lison and Bibauw(2017)}]{DBLP:conf/sigdial/LisonB17}
Pierre Lison and Serge Bibauw. 2017.
\newblock \href {https://doi.org/10.18653/v1/w17-5546} {Not all dialogues are
  created equal: Instance weighting for neural conversational models}.
\newblock In \emph{SIGdial Meeting on Discourse and Dialogue}, pages 384--394.
  Association for Computational Linguistics.

\bibitem[{Liu et~al.(2020)Liu, Chen, Chen, Lou, Chen, Zhou, and
  Zhang}]{RL-P2BOT-Liu2020}
Qian Liu, Yihong Chen, Bei Chen, Jian{-}Guang Lou, Zixuan Chen, Bin Zhou, and
  Dongmei Zhang. 2020.
\newblock \href {https://doi.org/10.18653/v1/2020.acl-main.131} {You impress
  me: Dialogue generation via mutual persona perception}.
\newblock In \emph{{ACL}}, pages 1417--1427.

\bibitem[{Pfeiffer et~al.(2021)Pfeiffer, Kamath, R{\"{u}}ckl{\'{e}}, Cho, and
  Gurevych}]{DBLP:conf/eacl/PfeifferKRCG21}
Jonas Pfeiffer, Aishwarya Kamath, Andreas R{\"{u}}ckl{\'{e}}, Kyunghyun Cho,
  and Iryna Gurevych. 2021.
\newblock \href {https://www.aclweb.org/anthology/2021.eacl-main.39/}
  {Adapterfusion: Non-destructive task composition for transfer learning}.
\newblock In \emph{{EACL}}, pages 487--503.

\bibitem[{Pfeiffer et~al.(2020)Pfeiffer, Vulic, Gurevych, and
  Ruder}]{DBLP:conf/emnlp/PfeifferVGR20}
Jonas Pfeiffer, Ivan Vulic, Iryna Gurevych, and Sebastian Ruder. 2020.
\newblock \href {https://doi.org/10.18653/v1/2020.emnlp-main.617} {{MAD-X:} an
  adapter-based framework for multi-task cross-lingual transfer}.
\newblock In \emph{{EMNLP}}, pages 7654--7673.

\bibitem[{Phang et~al.(2018)Phang, F{\'{e}}vry, and
  Bowman}]{DBLP:journals/corr/abs-1811-01088}
Jason Phang, Thibault F{\'{e}}vry, and Samuel~R. Bowman. 2018.
\newblock \href {http://arxiv.org/abs/1811.01088} {Sentence encoders on stilts:
  Supplementary training on intermediate labeled-data tasks}.
\newblock \emph{CoRR}, abs/1811.01088.

\bibitem[{Philip et~al.(2020)Philip, Berard, Gall{\'{e}}, and
  Besacier}]{DBLP:conf/emnlp/PhilipBGB20}
Jerin Philip, Alexandre Berard, Matthias Gall{\'{e}}, and Laurent Besacier.
  2020.
\newblock \href {https://doi.org/10.18653/v1/2020.emnlp-main.361} {Monolingual
  adapters for zero-shot neural machine translation}.
\newblock In \emph{{EMNLP}}, pages 4465--4470.

\bibitem[{Qian et~al.(2017)Qian, Huang, Zhao, Xu, and
  Zhu}]{Persona-QianQiao-2017}
Qiao Qian, Minlie Huang, Haizhou Zhao, Jingfang Xu, and Xiaoyan Zhu. 2017.
\newblock \href {http://arxiv.org/abs/1706.02861} {Assigning
  personality/identity to a chatting machine for coherent conversation
  generation}.
\newblock \emph{CoRR}, abs/1706.02861.

\bibitem[{Rashkin et~al.(2019)Rashkin, Smith, Li, and
  Boureau}]{DBLP:conf/acl/RashkinSLB19}
Hannah Rashkin, Eric~Michael Smith, Margaret Li, and Y{-}Lan Boureau. 2019.
\newblock \href {https://doi.org/10.18653/v1/p19-1534} {Towards empathetic
  open-domain conversation models: {A} new benchmark and dataset}.
\newblock In \emph{{ACL}}, pages 5370--5381.

\bibitem[{Ren et~al.(2018)Ren, Zeng, Yang, and
  Urtasun}]{DBLP:conf/icml/RenZYU18}
Mengye Ren, Wenyuan Zeng, Bin Yang, and Raquel Urtasun. 2018.
\newblock \href {http://proceedings.mlr.press/v80/ren18a.html} {Learning to
  reweight examples for robust deep learning}.
\newblock In \emph{{ICML}}, volume~80 of \emph{Proceedings of Machine Learning
  Research}, pages 4331--4340.

\bibitem[{Roller et~al.(2020)Roller, Dinan, Goyal, Ju, Williamson, Liu, Xu,
  Ott, Shuster, Smith, Boureau, and Weston}]{Blender-Roller-2020}
Stephen Roller, Emily Dinan, Naman Goyal, Da~Ju, Mary Williamson, Yinhan Liu,
  Jing Xu, Myle Ott, Kurt Shuster, Eric~Michael Smith, Y{-}Lan Boureau, and
  Jason Weston. 2020.
\newblock \href {http://arxiv.org/abs/2004.13637} {Recipes for building an
  open-domain chatbot}.
\newblock \emph{CoRR}, abs/2004.13637.

\bibitem[{Rust et~al.(2021)Rust, Pfeiffer, Vulic, Ruder, and
  Gurevych}]{DBLP:conf/acl/RustPVRG20}
Phillip Rust, Jonas Pfeiffer, Ivan Vulic, Sebastian Ruder, and Iryna Gurevych.
  2021.
\newblock \href {https://doi.org/10.18653/v1/2021.acl-long.243} {How good is
  your tokenizer? on the monolingual performance of multilingual language
  models}.
\newblock In \emph{{ACL/IJCNLP}}, pages 3118--3135.

\bibitem[{Sagi and Rokach(2018)}]{DBLP:journals/widm/SagiR18}
Omer Sagi and Lior Rokach. 2018.
\newblock \href {https://doi.org/10.1002/widm.1249} {Ensemble learning: {A}
  survey}.
\newblock \emph{Wiley Interdiscip. Rev. Data Min. Knowl. Discov.}, 8(4).

\bibitem[{See et~al.(2019)See, Roller, Kiela, and
  Weston}]{FilterSpecificity-See-2019}
Abigail See, Stephen Roller, Douwe Kiela, and Jason Weston. 2019.
\newblock \href {https://doi.org/10.18653/v1/n19-1170} {What makes a good
  conversation? how controllable attributes affect human judgments}.
\newblock In \emph{{NAACL-HLT}}, pages 1702--1723.

\bibitem[{Serban et~al.(2017{\natexlab{a}})Serban, Klinger, Tesauro,
  Talamadupula, Zhou, Bengio, and Courville}]{DBLP:conf/aaai/SerbanKTTZBC17}
Iulian~Vlad Serban, Tim Klinger, Gerald Tesauro, Kartik Talamadupula, Bowen
  Zhou, Yoshua Bengio, and Aaron~C. Courville. 2017{\natexlab{a}}.
\newblock \href {http://aaai.org/ocs/index.php/AAAI/AAAI17/paper/view/14571}
  {Multiresolution recurrent neural networks: An application to dialogue
  response generation}.
\newblock In \emph{{AAAI}}, pages 3288--3294. {AAAI} Press.

\bibitem[{Serban et~al.(2016)Serban, Sordoni, Bengio, Courville, and
  Pineau}]{HRED-Dialogue-Serban-2016}
Iulian~Vlad Serban, Alessandro Sordoni, Yoshua Bengio, Aaron~C. Courville, and
  Joelle Pineau. 2016.
\newblock \href
  {http://www.aaai.org/ocs/index.php/AAAI/AAAI16/paper/view/11957} {Building
  end-to-end dialogue systems using generative hierarchical neural network
  models}.
\newblock In \emph{{AAAI}}, pages 3776--3784.

\bibitem[{Serban et~al.(2017{\natexlab{b}})Serban, Sordoni, Lowe, Charlin,
  Pineau, Courville, and Bengio}]{VHRED-Serban-2017}
Iulian~Vlad Serban, Alessandro Sordoni, Ryan Lowe, Laurent Charlin, Joelle
  Pineau, Aaron~C. Courville, and Yoshua Bengio. 2017{\natexlab{b}}.
\newblock \href {http://aaai.org/ocs/index.php/AAAI/AAAI17/paper/view/14567} {A
  hierarchical latent variable encoder-decoder model for generating dialogues}.
\newblock In \emph{{AAAI}}, pages 3295--3301.

\bibitem[{Shang et~al.(2015)Shang, Lu, and Li}]{Seq2Seq-ShangLifeng-2015}
Lifeng Shang, Zhengdong Lu, and Hang Li. 2015.
\newblock \href {https://doi.org/10.3115/v1/p15-1152} {Neural responding
  machine for short-text conversation}.
\newblock In \emph{{ACL} {(1)}}, pages 1577--1586.

\bibitem[{Shang et~al.(2018)Shang, Fu, Peng, Feng, Zhao, and
  Yan}]{DBLP:conf/ijcai/ShangFPFZY18}
Mingyue Shang, Zhenxin Fu, Nanyun Peng, Yansong Feng, Dongyan Zhao, and Rui
  Yan. 2018.
\newblock \href {https://doi.org/10.24963/ijcai.2018/603} {Learning to converse
  with noisy data: Generation with calibration}.
\newblock In \emph{{IJCAI}}, pages 4338--4344.

\bibitem[{Shen et~al.(2021)Shen, Zhan, Shen, Chen, Zhao, and
  Zhu}]{DBLP:conf/cikm/ShenZSCZZ21}
Lei Shen, Haolan Zhan, Xin Shen, Hongshen Chen, Xiaofang Zhao, and Xiaodan Zhu.
  2021.
\newblock \href {https://doi.org/10.1145/3459637.3482352} {Identifying
  untrustworthy samples: Data filtering for open-domain dialogues with bayesian
  optimization}.
\newblock In \emph{{CIKM} '21: The 30th {ACM} International Conference on
  Information and Knowledge Management, Virtual Event, Queensland, Australia,
  November 1 - 5, 2021}, pages 1598--1608. {ACM}.

\bibitem[{Song et~al.(2020)Song, Wang, Zhang, Liu, and
  Liu}]{DBLP:conf/acl/SongWZLL20}
Haoyu Song, Yan Wang, Weinan Zhang, Xiaojiang Liu, and Ting Liu. 2020.
\newblock \href {https://doi.org/10.18653/v1/2020.acl-main.516} {Generate,
  delete and rewrite: {A} three-stage framework for improving persona
  consistency of dialogue generation}.
\newblock In \emph{{ACL}}, pages 5821--5831.

\bibitem[{Sordoni et~al.(2015)Sordoni, Galley, Auli, Brockett, Ji, Mitchell,
  Nie, Gao, and Dolan}]{Seq2Seq-Sordoni-2015}
Alessandro Sordoni, Michel Galley, Michael Auli, Chris Brockett, Yangfeng Ji,
  Margaret Mitchell, Jian{-}Yun Nie, Jianfeng Gao, and Bill Dolan. 2015.
\newblock \href {https://doi.org/10.3115/v1/n15-1020} {A neural network
  approach to context-sensitive generation of conversational responses}.
\newblock In \emph{{HLT-NAACL}}, pages 196--205.

\bibitem[{Sun et~al.(2021)Sun, Feng, Li, Liu, and Li}]{DBLP:conf/acl/SunFLLL20}
Bin Sun, Shaoxiong Feng, Yiwei Li, Jiamou Liu, and Kan Li. 2021.
\newblock \href {https://doi.org/10.18653/v1/2021.acl-long.437} {Generating
  relevant and coherent dialogue responses using self-separated conditional
  variational autoencoders}.
\newblock In \emph{{ACL}}, pages 5624--5637.

\bibitem[{Tao et~al.(2018)Tao, Gao, Shang, Wu, Zhao, and
  Yan}]{CMHAM-TaoChongyang-2018}
Chongyang Tao, Shen Gao, Mingyue Shang, Wei Wu, Dongyan Zhao, and Rui Yan.
  2018.
\newblock \href {https://doi.org/10.24963/ijcai.2018/614} {Get the point of my
  utterance! learning towards effective responses with multi-head attention
  mechanism}.
\newblock In \emph{{IJCAI}}, pages 4418--4424.

\bibitem[{Tiedemann(2009)}]{opensubtitles2009}
Jörg Tiedemann. 2009.
\newblock \href {https://doi.org/10.1075/cilt.309.19tie} {\emph{News from
  OPUS—A Collection of Multilingual Parallel Corpora with Tools and
  Interfaces}}.

\bibitem[{Vaswani et~al.(2017)Vaswani, Shazeer, Parmar, Uszkoreit, Jones,
  Gomez, Kaiser, and Polosukhin}]{Transformer-Vaswani-2017}
Ashish Vaswani, Noam Shazeer, Niki Parmar, Jakob Uszkoreit, Llion Jones,
  Aidan~N. Gomez, Lukasz Kaiser, and Illia Polosukhin. 2017.
\newblock \href
  {https://proceedings.neurips.cc/paper/2017/hash/3f5ee243547dee91fbd053c1c4a845aa-Abstract.html}
  {Attention is all you need}.
\newblock In \emph{{NIPS}}, pages 5998--6008.

\bibitem[{Vinyals and Le(2015)}]{NoisyData-Vinyals-2015}
Oriol Vinyals and Quoc~V. Le. 2015.
\newblock \href {http://arxiv.org/abs/1506.05869} {A neural conversational
  model}.
\newblock In \emph{{ICML} Deep Learning Workshop}.

\bibitem[{Wang et~al.(2020)Wang, Tang, Duan, Wei, Huang, Ji, Cao, Jiang, and
  Zhou}]{DBLP:journals/corr/abs-2002-01808}
Ruize Wang, Duyu Tang, Nan Duan, Zhongyu Wei, Xuanjing Huang, Jianshu Ji,
  Guihong Cao, Daxin Jiang, and Ming Zhou. 2020.
\newblock \href {http://arxiv.org/abs/2002.01808} {K-adapter: Infusing
  knowledge into pre-trained models with adapters}.
\newblock \emph{CoRR}, abs/2002.01808.

\bibitem[{Xing et~al.(2017)Xing, Wu, Wu, Liu, Huang, Zhou, and
  Ma}]{TopicAware-XingChen-2017}
Chen Xing, Wei Wu, Yu~Wu, Jie Liu, Yalou Huang, Ming Zhou, and Wei{-}Ying Ma.
  2017.
\newblock \href {http://aaai.org/ocs/index.php/AAAI/AAAI17/paper/view/14563}
  {Topic aware neural response generation}.
\newblock In \emph{{AAAI}}, pages 3351--3357.

\bibitem[{Xu et~al.(2018{\natexlab{a}})Xu, Ren, Lin, and
  Sun}]{DPGAN-XuJingjing-2018}
Jingjing Xu, Xuancheng Ren, Junyang Lin, and Xu~Sun. 2018{\natexlab{a}}.
\newblock \href {https://doi.org/10.18653/v1/d18-1428} {Diversity-promoting
  {GAN:} {A} cross-entropy based generative adversarial network for diversified
  text generation}.
\newblock In \emph{{EMNLP}}, pages 3940--3949.

\bibitem[{Xu et~al.(2018{\natexlab{b}})Xu, Dusek, Konstas, and
  Rieser}]{FilterCoherence-Xu-2018}
Xinnuo Xu, Ondrej Dusek, Ioannis Konstas, and Verena Rieser.
  2018{\natexlab{b}}.
\newblock \href {https://doi.org/10.18653/v1/d18-1432} {Better conversations by
  modeling, filtering, and optimizing for coherence and diversity}.
\newblock In \emph{{EMNLP}}, pages 3981--3991.

\bibitem[{Xu et~al.(2017)Xu, Liu, Wang, Sun, Wang, Wang, and
  Qi}]{GAN-GANAEL-Xu2017}
Zhen Xu, Bingquan Liu, Baoxun Wang, Chengjie Sun, Xiaolong Wang, Zhuoran Wang,
  and Chao Qi. 2017.
\newblock \href {https://doi.org/10.18653/v1/d17-1065} {Neural response
  generation via {GAN} with an approximate embedding layer}.
\newblock In \emph{{EMNLP}}, pages 617--626.

\bibitem[{Yu et~al.(2017)Yu, Zhang, Wang, and Yu}]{SeqGan-YuLantao-2017}
Lantao Yu, Weinan Zhang, Jun Wang, and Yong Yu. 2017.
\newblock \href {http://aaai.org/ocs/index.php/AAAI/AAAI17/paper/view/14344}
  {Seqgan: Sequence generative adversarial nets with policy gradient}.
\newblock In \emph{{AAAI}}, pages 2852--2858.

\bibitem[{Zhang et~al.(2018{\natexlab{a}})Zhang, Lan, Guo, Xu, and
  Cheng}]{RL-Seq2seqCo-Zhang2018}
Hainan Zhang, Yanyan Lan, Jiafeng Guo, Jun Xu, and Xueqi Cheng.
  2018{\natexlab{a}}.
\newblock \href {https://doi.org/10.24963/ijcai.2018/635} {Reinforcing
  coherence for sequence to sequence model in dialogue generation}.
\newblock In \emph{{IJCAI}}, pages 4567--4573.

\bibitem[{Zhang et~al.(2019)Zhang, Lan, Pang, Guo, and
  Cheng}]{DBLP:conf/acl/ZhangLPGC19}
Hainan Zhang, Yanyan Lan, Liang Pang, Jiafeng Guo, and Xueqi Cheng. 2019.
\newblock \href {https://doi.org/10.18653/v1/p19-1362} {Recosa: Detecting the
  relevant contexts with self-attention for multi-turn dialogue generation}.
\newblock In \emph{{ACL}}, pages 3721--3730.

\bibitem[{Zhang et~al.(2018{\natexlab{b}})Zhang, Dinan, Urbanek, Szlam, Kiela,
  and Weston}]{PersonaChat-facebook-2018}
Saizheng Zhang, Emily Dinan, Jack Urbanek, Arthur Szlam, Douwe Kiela, and Jason
  Weston. 2018{\natexlab{b}}.
\newblock \href {https://doi.org/10.18653/v1/P18-1205} {Personalizing dialogue
  agents: {I} have a dog, do you have pets too?}
\newblock In \emph{{ACL}}, pages 2204--2213.

\bibitem[{Zhang et~al.(2020)Zhang, Sun, Galley, Chen, Brockett, Gao, Gao, Liu,
  and Dolan}]{DialoGPT-Zhang-2020}
Yizhe Zhang, Siqi Sun, Michel Galley, Yen{-}Chun Chen, Chris Brockett, Xiang
  Gao, Jianfeng Gao, Jingjing Liu, and Bill Dolan. 2020.
\newblock \href {https://doi.org/10.18653/v1/2020.acl-demos.30} {{DIALOGPT} :
  Large-scale generative pre-training for conversational response generation}.
\newblock In \emph{{ACL}}, pages 270--278.

\bibitem[{Zhao et~al.(2017)Zhao, Zhao, and
  Esk{\'{e}}nazi}]{kgCVAE-ZhaoTiancheng-2017}
Tiancheng Zhao, Ran Zhao, and Maxine Esk{\'{e}}nazi. 2017.
\newblock \href {https://doi.org/10.18653/v1/P17-1061} {Learning
  discourse-level diversity for neural dialog models using conditional
  variational autoencoders}.
\newblock In \emph{{ACL} {(1)}}, pages 654--664.

\bibitem[{Zhou et~al.(2018)Zhou, Huang, Zhang, Zhu, and
  Liu}]{Emotional-Zhouhao-2018}
Hao Zhou, Minlie Huang, Tianyang Zhang, Xiaoyan Zhu, and Bing Liu. 2018.
\newblock \href
  {https://www.aaai.org/ocs/index.php/AAAI/AAAI18/paper/view/16455} {Emotional
  chatting machine: Emotional conversation generation with internal and
  external memory}.
\newblock In \emph{{AAAI-18}}, pages 730--739.

\end{thebibliography}
\bibliographystyle{acl_natbib}

\appendix 

\section{More Discussions}

\textbf{Q1:} Why do we use L1 to reflect the distance between $z^{m}$ and $z$? And can this distance represent the importance of adapters? \\
\textbf{Q2:} Why do we choose Specificity \citep{FilterSpecificity-See-2019} as one of the baselines? What is the relationship between \citet{FilterSpecificity-See-2019} and our work? \\
\textbf{A1:} (1) One is that L1 is better than other measurements in terms of computational efficiency. The other is that the distance calculated by L1 has higher discrimination than other measurements (e.g., L2), making the probability distribution (the coefficient $\lambda$) more concentrated. 
We have observed that the probability distribution based on L2 is relatively uniform, so it will make Equation \ref{eq:af_zandlambda} more like calculating the mean value. We have also tried mutual information (MI) as \citep{DBLP:conf/icml/GuanWZCH019}, but it would reduce the model performance. The results are shown in Table \ref{tb:distance_compare}. (2) The model is trained to capture and save more attribute-related features in each adapter layer, which are quite different from the features of the base model. Therefore, the distance of the input features (from the base model) and the output features (from each adapter) can reflect how well the corresponding features are learned. The larger the distance is, the more the model needs the knowledge of this attribute-related adapter. And we should pay more attention to these attribute-related features. \\\textbf{A2:} Although \citet{FilterSpecificity-See-2019} does not filter the dataset, it evaluates the sample quality with the proposed scoring methods. More importantly, it clearly demonstrates that different kinds of high-quality data are beneficial for the model to learn attribute-related features effectively, which also supports the motivation of our work. In our experiment, we use the scoring method \textit{Specificity} from it as one of the baselines because it can well characterize the specificity of the samples, and specific tokens are helpful to enhance the response quality.
\begin{table}[h]
    \centering
    \small
    \begin{tabular}{l | c c c  c c}
    \toprule
        Distance & Dist-1  & Dist-2   & KL-1 & KL-2  &BLEU  \\ \midrule
        MI & 0.0412 & 0.1451  & 0.92  & 0.82  & 0.379  \\
        L2 & 0.0431 & 0.1520  & 0.90  & 0.77  & \textbf{0.384}  \\
        L1 (Ours) & \textbf{0.0434}  & \textbf{0.1522}  & \textbf{0.88}  & \textbf{0.75} & 0.383 \\
    \bottomrule
    \end{tabular}
    \caption{Comparison of different distance measurements for Adaptive Fusion (AF).}
    \label{tb:distance_compare}
\end{table}

\section{Which Samples Can Be Considered High-Quality?}
\label{appendix:a}

We utilize the Specificity \citep{FilterSpecificity-See-2019} and Consistency \citep{FilterConsistency-Akama-2020} scoring methods to evaluate the samples of DailyDialog, and obtain four sets of samples, shown in Figure~\ref{fig.multi_indicators}. We use these sets to train the Transformer-based dialogue model. The results are shown in Table~\ref{appendix tb:analysis-of-figure1}. ``Filtering-Con'', ``Filtering-Spe'', ``Intersection'', and ``Union'' represent the models trained on the blue+red parts, the blue+orange parts, the blue part, and the blue+red+orange parts in Figure~\ref{fig.multi_indicators}, respectively. 
From Table~\ref{appendix tb:analysis-of-figure1}, we can find that the performances of models trained on the blue+red parts and the blue+orange parts are both better than that of model trained only on the blue part. Besides, the model trained on the blue+red+orange parts obtains the best performance than others. These results illustrate that the samples with low scores in one scoring method but high scores in other scoring methods are still good samples for the model training.

\section{Algorithm of Proposed Framework}
\label{appendix:algorithm}
The full training details of MAE are shown in Algorithm \ref{appendix algorithm}.

\renewcommand{\algorithmicrequire}{\textbf{Input:}}
\renewcommand{\algorithmicensure}{\textbf{Output:}}
\begin{algorithm}[h]
    \caption{MAE}\label{alg:MAE}
    \begin{algorithmic}[1]
    \REQUIRE
    $\mathcal{D}$, $\mathcal{D}^{v}$ and $\mathcal{D}^{t}$: the raw training, validation and test dataset;\\
    $\mathcal{S}=\{\mathcal{S}_1, \mathcal{S}_2, \ldots, \mathcal{S}_M\}$ : the scoring methods;\\
    $\theta$ : the parameters of the base model;\\
    $\phi=\{\phi_{1},\phi_{2},\ldots,\phi_{M}\}$: the parameters of adapters;\\
    $Fusion\_flag$ : the flag used to choose AF or PF.
    \ENSURE
    $\theta^*$ and $\phi^*$ : the learned base model and adapters.
    \STATE \% View-specific collection.
    \FOR{$m=1$ to $M$}
        \STATE data\_scores $\leftarrow$ calculate\_data\_scores($\mathcal{D}$, $\mathcal{S}_m$)
        \STATE index\_list $\leftarrow$ sort(data\_scores)
        \STATE $\mathcal{D}_m$ $\leftarrow$ extract\_top\_data($\mathcal{D}$, index\_list)
    \ENDFOR
    \STATE \% Pre-train the base model $\theta$.
    \REPEAT
        \STATE optimize $\theta$ by minimizing $\mathcal{L}_{nll}(\theta)$ on $\mathcal{D}$ using Eq.~\eqref{eq:loss}
        \STATE evaluate $\theta$ on $\mathcal{D}^v$
    \UNTIL{convergence}
    \STATE \% Fine-Tune adapters with fixed $\theta^*$ and fusion.
    \IF{Fusion\_flag is AF}
        \REPEAT
            \STATE optimize $\phi_{1},\phi_{2},\ldots,\phi_{M}$ in parallel by minimizing $\mathcal{L}_{nll}(\phi^{m})$ on its corresponding $\mathcal{D}_m$
            \STATE evaluate $\theta^{*}+\phi^{m}$ on $\mathcal{D}^v$
        \UNTIL{convergence}
        \STATE fuse $\phi^{*}_1,\phi^{*}_2,\ldots,\phi^{*}_M$ using Eq.~\eqref{eq:af_zandlambda} on $\mathcal{D}^t$
    \ELSE 
    \STATE \% Fusion\_flag is PF
    \FOR{$m=1$ to $M$}
        \STATE $m$ $\leftarrow$ randomly\_pop($1, 2,\ldots, M$)
        \REPEAT
            \STATE \% Fusion during training.
            \STATE optimize $\phi^{m}$ by minimizing $\mathcal{L}(\phi^{m})$ on $\mathcal{D}_m$ using Eq.~\eqref{eq:pf}
            \STATE evaluate $\theta^{*}+\phi^{1}+\ldots+\phi^{m}$ on $\mathcal{D}^v$
        \UNTIL{convergence}
    \ENDFOR
    \ENDIF
    \RETURN learned base model $\theta^*$ and adapters $\phi^*=(\phi^{*}_1,\phi^{*}_2,\ldots,\phi^{*}_M)$
    \end{algorithmic}
    \label{appendix algorithm}
\end{algorithm}

\section{Datasets}
\label{details for dataset}
Two public dialogue datasets are employed in our experiments:
\textbf{DailyDialog}, which contains conversations that are similar to human daily communication \citep{dailydialog2017}, and \textbf{OpenSubtitles}, which consists of large-scale dialogues converted from movie subtitles \citep{opensubtitles2009}. 

Table~\ref{appendix tab:data_statistics} provides the statistics of both datasets after data preprocessing.

\section{Details for Scoring Methods}
\label{details for scoring methods}
Here are the details of three high-quality automatic scoring methods compared by the experiments:

\begin{itemize}
\item     \textbf{Consistency} \citep{FilterConsistency-Akama-2020}: 
A joint score:
\begin{equation}
S_{C+R}(q, r) = \alpha S_{C} + \beta S_{R}
\end{equation}
that consists of two parts: connectivity $S_{C}$ and content relatedness $S_R$. The $\alpha$ and $\beta$ are hyper-parameters that weigh the two parts, and are fixed as the means of all $S_C$ and $S_R$, respectively.
    The $S_{C}$ is evaluated by the co-occurrence of key-phrases ($p\in q$, $h\in r$):
    \begin{equation}
    \small
    \label{eq:sc}
         S_C = 
         \sum_{(p,h)} \frac{\max(nPMI(p,h),0)\cdot |p| \cdot |h|} {|q| \cdot |r|},
    \end{equation}
    where $|\cdot|$ means the number of words in the phrase or utterance, and the $nPMI$ represents the normalized pointwise mutual information \citep{nPMI-Bouma-2009}. In addition, $S_{R}$ is evaluated by the cosine of the context and its response:
    \begin{equation}
    \small
        S_R = \max(cos(q_{emb}, r_{emb}),0)
    \end{equation}
    The $q_{emb}$ and the $r_{emb}$ are vector representations of the query and response. This scoring method can reflect the consistency of a dialogue pair. 
    
\item    \textbf{Entropy\_Src} \citep{FilterEntropy-Csaky-2019}: 
This score is the entropy of a response utterance:
    \begin{equation}
    \small
        H_{src}(r|D) = -\sum_{(q_i, r)\in D}p(q_i|r)\log{p(q_i|r)},
    \end{equation}
    where $r$ represents the response, $D$ represents the dialogue dataset, and $q_i$ means a query of $r$ in $D$. The $p(q_i|r)$ means the probability that the query is $q_i$ while the response is $r$. This scoring method will filter the dialogue pair with "many-to-one" problem, so that it will alleviate the phenomenon of general response.
    
    \item \textbf{Specificity} \citep{FilterSpecificity-See-2019}:
    \begin{equation}
    \small
        NIDF(t) = \frac{idf(t)-min\_idf}{max\_idf-min\_idf},
    \end{equation}
    where the $t$ is a token of the response, and the $idf(t)=\log(\frac{R}{R_t})$. $R$ is the number of responses in the dataset, and $R_t$ is the number of those responses that contain $t$. The mean Normalized Inverse Document Frequency (NIDF) of all tokens in an utterance is utilized to represent the specificity of it. This scoring method can identify whether the response contains specific tokens.
\end{itemize}

\begin{table}[t]
    \centering
    \small
    \begin{tabular}{l l l l l l}
    \toprule
        & Dist-1 & Dist-2 & KL-1 & KL-2 & BLEU \\ \midrule
        Filtering-Con & 0.0088 & 0.0306 & 2.74 & 2.01 & 0.314 \\
        Filtering-Spe & 0.0069 & 0.0321 & 2.63 & 2.86 & 0.190 \\
        Intersection & 0.0042 & 0.0191 & 3.59 & 3.27 & 0.185 \\ 
        Union & 0.0263 & 0.0982 & 1.38 & 1.08 & 0.332 \\
    \bottomrule
    \end{tabular}
    \caption{Results of the model trained on different sets of samples in Figure \ref{fig.multi_indicators}.}
    \label{appendix tb:analysis-of-figure1}
\end{table}

\begin{table}[t]
\centering
\small
\begin{tabular}{lllll}
\toprule
    Datasets      & Vocab  & Train & Valid & Test \\ \midrule
    DailyDialog         & 17,930 & 68k & 6.8k & 6.8k \\
    OpenSubtitles       & 21,177 & 200k & 20k & 10k   \\
\bottomrule
\end{tabular}
\caption{Statistics for DailyDialog and OpenSubtitles.}
\label{appendix tab:data_statistics}
\end{table}

\section{Training Details}
\label{training details}

\begin{table}[t]
\centering
\small
\begin{tabular}{ll}
\toprule
    Name            & Value \\ \midrule
    Hidden size               & 512 \\
    Number of hidden layers   & 6   \\
    Number of attention heads & 8   \\ 
    Feed-forward units       & 2048 \\
    Dimension of query,key,value & 64 \\
    Embedding size            & 512 \\
    Label smoothing           & 0.1 \\
    Layer dropout             & 0.1 \\
    Relu dropout              & 0.1 \\
    Attention dropout         & 0.1 \\
\bottomrule
\end{tabular}
\caption{Transformer hyperparameters.}
\label{tab:trsfm_parameters}
\end{table}

The hyparameters of our Transformer based dialogue model is shown in Table~\ref{tab:trsfm_parameters}. We use the Adam optimizer \citep{DBLP:journals/corr/KingmaB14} and employ \textit{warm up} trick to adjust the learning rate during training with the $warm\_up\_steps$ set as 32,000, which is computed as:
\begin{equation}
    lr = \frac{2\times \min(\frac{1}{\sqrt{n\_steps}}, \frac{n\_steps}{\sqrt{warm\_up\_steps^3}})}{\sqrt{d\_model}},
\end{equation} 
where $lr$ is the learning rate at the $n\_steps$ of training. For our method, we set the units of down-projection and up-projection feed-forward networks of each adapter as 64 and 512, respectively. In inference stage, the Beam Search is employed, and the beam size is set as 5.

Details for the metrics we employ for both automatic and human evaluations:

\begin{itemize}
\item \textbf{Dist-\{1,2\}} (distinct) \citep{MMI-LiJiwei-2016} is a widely used metric that reflects the lexical diversity of the generated responses by calculating the proportion of unique unigrams/bigrams.

\item \textbf{KL-\{1,2\}} (KL divergence) \citep{FilterEntropy-Csaky-2019} measures the distribution distance between the generated and the ground-truth response sets to reflect how well a model can approximate the ground-truth unigrams/bigrams distribution. 

\item \textbf{BLEU} \citep{DBLP:conf/wmt/ChenC14} measures n-gram overlap between the generated and the ground-truth responses. 
\item \textbf{Coherence} (Xu et al., 2018b) measures the cosine similarity between pairs of input and response. 
\item \textbf{H-\{1,2\}} (word entropy) (Serban et al., 2017b) measures the unigrams/bigrams' non-genericness of responses by $H=-\frac{1}{|U|} \sum_{w \in U} \log _{2} p(w)$, where $p(w)$ is calculated based on frequency observed in the training data.
\item \textbf{Informativeness} reflects how much the information related to the query is contained in the generated response.
\item \textbf{Relevance} reflects how likely the generated response is relevant to its query.
\item \textbf{Fluency} reflects how likely the generated response comes from human. 
\item \textbf{Consistency} reflects how likely the generated response is coherent to its query, roughly the same as \textbf{Relevance}.
\item \textbf{Specificity} reflects how much the generated response is good at word usage.
\end{itemize}

\section{Detailed Kappa Results for Human Evaluations}

\label{kappa}
\begin{table}[ht]
    \centering
    \small
    \renewcommand\tabcolsep{2.0pt}
    \begin{tabular}{lccc}
     \toprule
      vs. Models & Informativeness & Relevance & Fluency \\ \midrule
      Transformer & 0.537/0.515  & 0.459/0.472  & 0.520/0.599  \\ 
 Filtering-Con  &        0.652/0.681  & 0.586/0.631  & 0.474/0.391  \\ 
 Filtering-Ent & 0.450/0.604  & 0.555/0.667  & 0.458/0.467  \\
 Filtering-Spe & 0.362/0.534  & 0.620/0.664  & 0.527/0.560  \\ 
 Weighting-Ent & 0.511/0.631  & 0.479/0.652  & 0.569/0.522  \\ 
\midrule
      Transformer & 0.630/0.688  & 0.734/0.604  & 0.715/0.627  \\
 Filtering-Con  &0.690/0.575  & 0.659/0.498  & 0.537/0.501  \\
 Filtering-Ent&0.605/0.615  & 0.745/0.650  & 0.571/0.548 \\
 Filtering-Spe&0.703/0.595  & 0.574/0.541  & 0.606/0.640  \\
 Weighting-Ent &0.759/0.632  & 0.686/0.624  & 0.564/0.599  \\
    \bottomrule
    \end{tabular}
    \caption{Fleiss's Kappa for human evaluations on DailyDialog (Top) and OpenSubtitles (Bottom). A/B in each table cell refer to the results of MAE-AF/MAE-PF, respectively.}
    \label{tb:human eval kappa}
\end{table}

Table \ref{tb:human eval kappa} shows the detailed results of Fleiss' Kappa \citep{fleisskappa/measuring} for human evaluations.


\section{Study on the Order of Progressive Fusion}
\label{appendix:order}
\begin{table}[htb]
    \centering
    \small
    \renewcommand\tabcolsep{2.5pt}
    \begin{tabular}{l l l l l l l l}
    \toprule
        Models & Dist-1  & Dist-2   & KL-1 & KL-2  &BLEU  \\ \midrule
        Transformer & 0.0157  & 0.0410 & 2.37  & 1.72  & 0.336 \\  
        MAE-AF & 0.0204 & 0.0660 & 1.83  & 1.55 & 0.335  \\
        MAE-PF-CES & 0.0217  & 0.0676  & 1.80  & 1.46 & 0.339   \\
        MAE-PF-CSE & 0.0210  & 0.0681  & 1.72  & 1.55 & 0.328  \\ 
        MAE-PF-SEC & 0.0224  & 0.0723  & 1.71  & 1.38 & 0.337  \\ 
        MAE-PF-SCE & 0.0218  & 0.0702  & 1.66  & 1.45 & 0.332 \\ 
        MAE-PF-ESC & 0.0224  & 0.0736  & 1.63  & 1.39 & 0.336  \\
        MAE-PF-ECS & 0.0236  & 0.0762  & 1.63  & 1.32 & 0.340 \\ 
    \bottomrule
    \end{tabular}
    \caption{Results of MAE-PF with different fusion orders.}
    \label{tb:order}
\end{table}

In order to study the influence of progressive fusion order on the model performance, we consider all possible sequences and conduct corresponding experiments. Table \ref{tb:order} shows that no matter what kind of progressive training order, the results are significantly better than the baseline model. And the distinct and KL divergence are also improved compared with the adaptive fusion (AF) strategy. This proves that all the different learning order for PF can get great improvement.

\section{More Results for Robustness Analysis of the Selection Ratio}
\label{appendix:volume all}

To investigate the effect of the selection ratio on the performance of filtering methods, we expand the test with two different proportions on both two datasets. The experimental results are summarized in Table~\ref{appendix tb:volume all}. Compared with the non-filtered base transformer model, the three 80\% filtered models make slight improvement of OpenSubtitles on the whole, while decrease in some degree on DailyDialog. The reason for this difference lies in the different amounts of data between the two datasets. This shows that data filtering is not a good training optimization method for small datasets. While for 50\% filtering, the results of both datasets decrease, markedly on DailyDialog. It is illustrated that simply filtering the dataset is prone to harm the training because the dropped data may contain a lot of useful knowledge in other dialogue attributes. As a result, a better way is knowledge enhancement by the adapter rather than filtering for the use of scoring methods in dialogue generation.

Table~\ref{ap:robust} shows the detailed results of Figure~\ref{fig:a4}.

\begin{table}[h]
    \centering
    \small
    \renewcommand\tabcolsep{2.5pt}
    \begin{tabular}{l l l l l l l l}
    \toprule
        Models & Dist-1  & Dist-2   & KL-1 & KL-2  &BLEU \\ \midrule
        Transformer & 0.0216 & 0.0728 & 1.67  & 1.66 & 0.292  \\
        MAE-Con (80$\%$) & 0.0370   & 0.1308  & 1.02  & 0.85  & 0.373 \\ 
        MAE-Con (70$\%$) & 0.0368   & 0.1287  & 1.03  & 0.86  & 0.374 \\ 
        MAE-Con (60$\%$) & 0.0355   & 0.1203  & 1.06  & 0.93  & 0.362    \\ 
        MAE-Con (50$\%$) & 0.0356   & 0.1239  & 1.07  & 0.97  & 0.362   \\ 
        MAE-Con (40$\%$) & 0.0358   & 0.1234  & 1.08  & 0.98  & 0.358   \\ 
        MAE-Con (30$\%$) & 0.0354   & 0.1191  & 1.12  & 1.00  & 0.356   \\ 
        MAE-Con (20$\%$) & 0.0348   & 0.1164  & 1.15  & 1.06  & 0.356   \\ 
        MAE-Con (10$\%$) & 0.0314   & 0.1048  & 1.34   & 1.27  & 0.334   \\ 
    \bottomrule
    \end{tabular}
    \caption{Detailed results of Figure~\ref{fig:a4}.}
    \label{ap:robust}
\end{table}

\begin{table*}[!t]
    \centering
    \small
    \begin{tabular}{l|l l l l l| l l l l l}
    \toprule
        Models & Dist-1  & Dist-2  & KL-1 & KL-2 &BLEU  & Dist-1  & Dist-2  & KL-1 & KL-2 &BLEU  \\ \midrule
        Transformer & 0.0216 & 0.0728 & 1.67  & 1.66 & 0.292 & 0.0157  & 0.0410 & 2.37  & 1.72  & 0.336 \\
        Filtering-Con (80$\%$) & 0.0225  & 0.0752  & 1.62  & 1.08 & 0.330  & 0.0189  & 0.0569 & 2.01  & 1.70 & 0.322 \\
        Filtering-Con (50$\%$) & 0.0088  & 0.0306 & 2.74  & 2.01 & 0.314 & 0.0160  & 0.0544 & 1.87  & 1.46 & 0.321 \\
        Filtering-Ent (80$\%$) & 0.0166  & 0.0462  & 2.16  & 1.77  & 0.307  & 0.0156  & 0.0429 & 2.42  & 1.73  & 0.337 \\ 
        Filtering-Ent (50$\%$) & 0.0096  & 0.0255 & 2.63  & 2.17 & 0.315 & 0.0102  & 0.0291  & 2.72  & 1.52 & 0.366 \\ 
        Filtering-Spe (80$\%$) & 0.0132  & 0.0465  & 2.20  & 2.24  & 0.244 & 0.0150 &0.0439   & 1.93  & 1.82  & 0.311 \\
        Filtering-Spe (50$\%$) & 0.0061  & 0.0210 & 3.62  & 3.17  & 0.199  & 0.0089 & 0.0235 & 3.02  & 1.83 & 0.348 \\ 
        MAE-AF (80$\%$)& 0.0383  & 0.1370  & 0.97 & 0.85 & 0.375  & 0.0194 & 0.0585  & 1.91  & 1.57 & 0.328   \\
        MAE-AF (50$\%$)& 0.0434  & 0.1522  & 0.88  & 0.75 & 0.383  & 0.0204  & 0.0660 & 1.83  & 1.55 & 0.335  \\
        MAE-PF (80$\%$) & 0.0403 & 0.1361 & 0.94  & 0.68  & 0.382 & 0.0204  & 0.0644 & 1.77  & 1.58 & 0.326 \\
        MAE-PF (50$\%$) & 0.0463 & 0.1511 & 0.94  & 0.68  & 0.392 & 0.0217  & 0.0676 & 1.80  & 1.46 & 0.339 \\
    \bottomrule
    \end{tabular}
    \caption{Impact of the selection ratio on the model performance on DailyDialog (Left) and OpenSubtitles (Right).}
    \label{appendix tb:volume all}
\end{table*}

\end{document}